\documentclass[10pt,twocolumn,letterpaper]{article}

\usepackage[pagenumbers]{cvpr} 

\usepackage{graphicx}
\usepackage{amsmath}
\usepackage{amssymb}
\usepackage{booktabs}
\usepackage{epsfig}
\usepackage{color, colortbl, xcolor}
 
\usepackage[normalem]{ulem} 
\usepackage{cuted}%
\usepackage{caption}
\usepackage{comment}
\usepackage{enumitem}
\usepackage[accsupp]{axessibility} 

\usepackage[pagebackref,breaklinks,colorlinks]{hyperref}

\definecolor{pastelred}{rgb}{1.0, 0.41, 0.38} 
\definecolor{ao(english)}{rgb}{0.0, 0.5, 0.0}

\def\cvprPaperID{2077} 
\def\confName{CVPR}
\def\confYear{2022}

\begin{document}
 
\title{Assembly101: A Large-Scale Multi-View Video Dataset \\ 
for Understanding Procedural Activities} 

\author{Fadime Sener$^{\dagger}$ \;\; Dibyadip Chatterjee$^{\ddagger}$ \;\; Daniel Shelepov$^{\dagger}$ \;\; Kun He$^{\dagger}$ \;\; Dipika Singhania$^{\ddagger}$ \\ Robert Wang$^{\dagger}$ \;\; Angela Yao$^{\ddagger}$\vspace{10pt}\\ 
		$^\dagger$ Meta Reality Labs Research \;\; 
		$^\ddagger$ National University of Singapore\\ 
  {\tt\small \{famesener,dsh,kunhe,rywang\}@fb.com} \;\;
 {\tt\small \{dibyadip,dipika16,ayao\}@comp.nus.edu.sg}\\ 
		\textcolor{blue}{\url{https://assembly-101.github.io/}\vspace{-6pt}}
}

\maketitle

\begin{abstract}
\vspace{-3mm}
Assembly101 is a new procedural activity dataset featuring 4321 videos of people assembling and disassembling 101 ``take-apart'' toy vehicles. Participants work without fixed instructions, and the sequences feature rich and natural variations in action ordering, mistakes, and corrections. Assembly101 is the first multi-view action dataset, with simultaneous static (8) and egocentric (4) recordings. Sequences are annotated with more than 100K coarse and 1M fine-grained action segments, and 18M 3D hand poses. 

We benchmark on three action understanding tasks: recognition, anticipation and temporal segmentation. Additionally, we propose a novel task of detecting mistakes. The unique recording format and rich set of annotations allow us to investigate generalization to new toys, cross-view transfer, long-tailed distributions, and pose vs. appearance. We envision that Assembly101 will serve as a new challenge to investigate various activity understanding problems.
\end{abstract}

\begin{figure*}[t!]
\centering 
\includegraphics[width=0.97\textwidth]{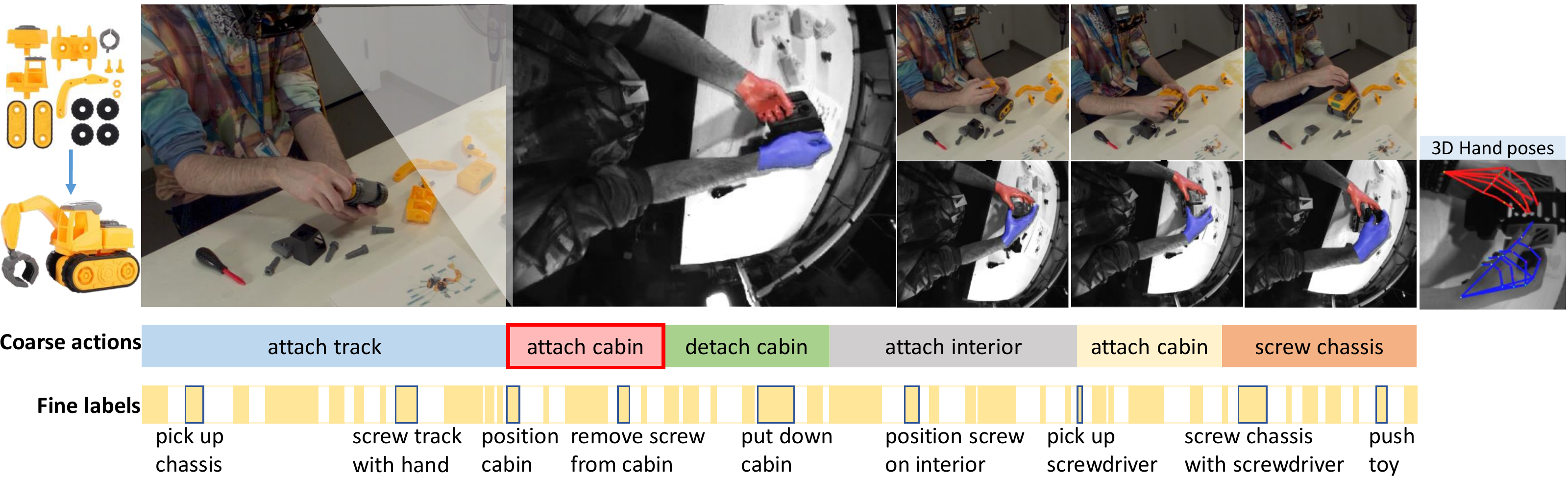}
\vspace{-2mm}
\captionof{figure}{Assembly101 includes synchronized static multi-view and egocentric recordings of participants assembling and disassembling take-apart toys. Sequences are annotated with fine-grained and coarse actions, 3D hand poses, participants' skill levels, and mistakes on coarse segments (\eg ``attach cabin'' highlighted in red). } 
\label{fig:teaser}
\end{figure*}

\vspace{-3mm}
\section{Introduction}
Assembly and disassembly tasks, like putting together a piece of furniture, or taking apart a home appliance for repair, are common to everyday living. We often rely on paper manuals or online instructional videos to guide us through these tasks. The next generation of smart assistants, together with augmented reality~(AR) hardware, can help us in a more embodied setting. Intelligent systems that jointly consider instructions or goals \emph{and} real-world observations can greatly advance AR applications. Mock-ups and proof-of-concepts already exist for cooking~\cite{googleglass}, monitoring worker safety~\cite{colombo2019deep}, visiting museums~\cite{farinella2019vedi}, and learning surgical procedures~\cite{beyer2016virtual}. To that end, the interest in action understanding tasks such as recognition, anticipation, and temporal segmentation has grown, especially for egocentric views~\cite{Damen2018EPICKITCHENS,Ego4D2021,ragusa2021meccano}. 

In looking at the benchmarks used in action understanding, there are datasets of short clips~\cite{soomro2012ucf101,kay2017kinetics,goyal2017something}, datasets with longer sequences from movies~\cite{zhu2015aligning,gu2018ava} and scripted actions~\cite{sigurdsson2016hollywood,sigurdsson2018actor,stein2013combining}, with particular focus on the cooking domain~\cite{fathi2011learning,rohrbach2012database,regneri2013grounding,stein2013combining,kuehne2014language,Damen2018EPICKITCHENS,zhou2018towards,sener2019zero}. Most related to our work are instructional video datasets~\cite{zhukov2019cross,tang2019coin,zhou2018towards}. But these instructional videos are curated from online sources; they are produced, have multiple shots, and primarily target multi-modal (vision + NLP) learning~\cite{zhukov2019cross,sener2019zero,zhou2018towards}. Few datasets focus on goal-oriented, multi-step activities outside the kitchen domain and are otherwise small-scale~\cite{ragusa2021meccano,EgocentricTent,ben2021ikea} or limited in task or sequence diversity~\cite{alayrac2016unsupervised,tang2019coin}. 

We introduce Assembly101: 362 unique sequences of people assembling and disassembling 101 ``take-apart'' toy vehicles (see Figs.~\ref{fig:teaser},~\ref{fig:alltoys}). The dataset features recordings from 8 static and 4 egocentric viewpoints, with 4321 sequences totalling 513 hours of footage. Assembly101 is annotated with more than 1M action segments, spanning 1380 fine-grained and 202 coarse action classes. We benchmark on four tasks: \emph{action recognition} and \emph{anticipation} centered around hand-object interactions, \emph{temporal action segmentation} and our newly proposed \emph{mistake detection} task dedicated to investigating sequence understanding in assembly activities. Assembly101 features three novel aspects currently under-represented in existing video benchmarks: 

\setlist{nolistsep}
\begin{itemize}[noitemsep]
\item \textbf{Goal-oriented free-style procedures:} Existing datasets feature multi-step activities following a strictly ordered recipe~\cite{zhukov2019cross,zhou2018towards,miech19howto100m,sener2019zero} or script~\cite{ragusa2021meccano,GTEAGaze,fathi2011learning,stein2013combining,sigurdsson2016hollywood}. Assembly101 depicts non-scripted, goal-oriented activities. 

\item \textbf{Rich sequence variation:} Participants vary in skill level, and recordings feature realistic variations in action ordering, mistakes, and corrections. Unlike existing skill assessment datasets~\cite{doughty2019pros,gao2014jhu,pirsiavash2014assessing,parmar2019action}, which have only skill scores, we annotate specific mistakes and participant skill levels. 

\item \textbf{Synchronized static and egocentric viewpoints:} This unique multi-view setting gives privileged static information currently missing from egocentric datasets. It also allows for investigating hand-object interactions with full 3D understanding and domain-transfer between different viewpoints. 
\end{itemize}

\section{A Comparison of Action Datasets}
Assembly101 can be characterized by its (1) multi-step content, (2) multi-view recordings and (3) action understanding tasks. We make a coarse comparison to related datasets based on this taxonomy. 

\subsection{Content: Multi-step activities} 
Multi-step activities are best exemplified in cooking and instructional videos, so the majority of datasets in this area are curated from online video platforms,~\eg YouTube Instructional~\cite{alayrac2016unsupervised}, What’s Cooking~\cite{malmaud2015s}, YoucookII~\cite{zhou2018towards}, CrossTask~\cite{zhukov2019cross}, COIN~\cite{tang2019coin} and HowTo100M~\cite{miech2019howto100m}. Using YouTube videos is appealing due to the sheer amount and variety. However, these videos often do not suit an AR setting due to their ``produced'' nature, \eg mixed viewpoints, fast-forwarding, unrelated narrations, etc. Additionally, the majority of these datasets are from the kitchen domain and are primarily composed for studying multi-modal learning in vision and natural language~\cite{malmaud2015s,zhou2018towards,zhukov2019cross}. 

Recorded datasets, \eg Breakfast~\cite{kuehne2014language}, GTEA~\cite{fathi2011learning}, 50Salads~\cite{stein2013combining} are major contributors to the study of multi-step activities~\cite{farha2019ms,sener2018unsupervised,farha2020long}. However, they are either small~\cite{fathi2011learning,stein2013combining} or have little ordering variations~\cite{kuehne2014language}. Assembly tasks are a new domain explored in some datasets~\cite{ragusa2021meccano,ben2021ikea}, but their limited scale is less ideal for deep learning.

\subsection{Viewpoint: Egocentric \& multi-view}
\noindent\textbf{Egocentric} data offers a unique viewpoint for human activities and is particularly important for wearables, \eg AR glasses. Small-scale datasets include~\cite{pirsiavash2012detecting,fathi2011learning,ragusa2021meccano,EgocentricTent}. Large-scale efforts include EPIC-KITCHENS~\cite{Damen2018EPICKITCHENS,Damen2020RESCALING} and the recent Ego4D~\cite{Ego4D2021}, which expands beyond the kitchen to a wide variety of daily activities. In contrast to these datasets, Assembly101 features both egocentric and third-person views, offering simultaneous privileged information from the outside-in as well as multi-view egocentric data for 3D action recognition. 
 
\noindent\textbf{Multi-view} fixed-camera datasets include IKEA~\cite{ben2021ikea} and Breakfast~\cite{kuehne2014language}. We feature a synchronized egocentric stream that allows studying the domain gap between fixed and egocentric views. Moreover, the egocentric head pose is tracked relative to the fixed views, enabling geometric reasoning between the viewpoints. Although Charades-EGO~\cite{sigurdsson2018actor} also has both an egocentric and a third-person view of people performing scripted activities, the views are taken asynchronously, \ie independent recording instances. 

\subsection{Task}
\noindent\textbf{Action recognition:} We focus on fine-grained actions lasting a few seconds within the context of longer activity sequences. This is in contrast to classifying short isolated clips, such as in Kinetics~\cite{kay2017kinetics} and Something-Something~\cite{goyal2017something}. Our task is more similar to EPIC-KITCHENS~\cite{Damen2018EPICKITCHENS} and Charades~\cite{sigurdsson2016hollywood,sigurdsson2018actor}, which feature fine-grained segments taken from longer daily activity videos with challenging long-tail distributions. 
 
\noindent\textbf{Anticipating actions} before they occur is a recently introduced task popularized by EPIC-KITCHENS~\cite{Damen2018EPICKITCHENS} and Breakfast~\cite{kuehne2014language}. A notable difference between these two is the label granularity and hence the anticipation horizon. Anticipation methods for EPIC predict fine-grained actions with a short, few-second long horizon, while Breakfast aims to predict multiple coarse actions with minutes-long horizons. As Assembly101 features multi-granular labels, it can be used for both short- and long-horizon anticipation. 

\noindent\textbf{Temporal action segmentation} datasets like GTEA~\cite{fathi2011learning} and 50Salads~\cite{stein2013combining} are small-scale datasets (28 and 50 videos respectively). Breakfast~\cite{kuehne2014language} is limited in temporal variation, making it less ideal for studying sequencing and ordering as a problem. The assembly actions in our dataset feature repetitions, large deviations in ordering and also require modelling longer-range information.
 
\noindent\textbf{Hand-object interactions} from egocentric views are studied in two new datasets, FPHA~\cite{garcia2018first} and H2O~\cite{kwon2021h2o}. Unlike EPIC, FPHA and H2O provide 3D pose of one or both hands and 6D pose of the manipulated objects. Recognition from pose is particularly important when the amount of visual data given to the system is limited, \eg due to privacy concerns. Assembly101 currently offers 3D hand poses for each frame. It offers a much larger set of fine-grained hand-object interactions compared to FPHA and H2O. 

\noindent\textbf{Detecting mistakes} and missed actions by wearable devices could greatly improve wearer's safety. Anomaly detection in surveillance videos~\cite{sultani2018real} and skill assessment~\cite{pan2019action,zia2016automated,doughty2019pros,gao2014jhu} are active research areas, but to the best of our knowledge, detecting mistakes in procedural activities has not been previously studied. The coarse action segments of our assembly sequences are annotated with mistake labels. Closest to our work is~\cite{soran2015generating} on forgotten actions. 

\begin{figure}[t]
\centering 
\includegraphics[width=0.87\linewidth]{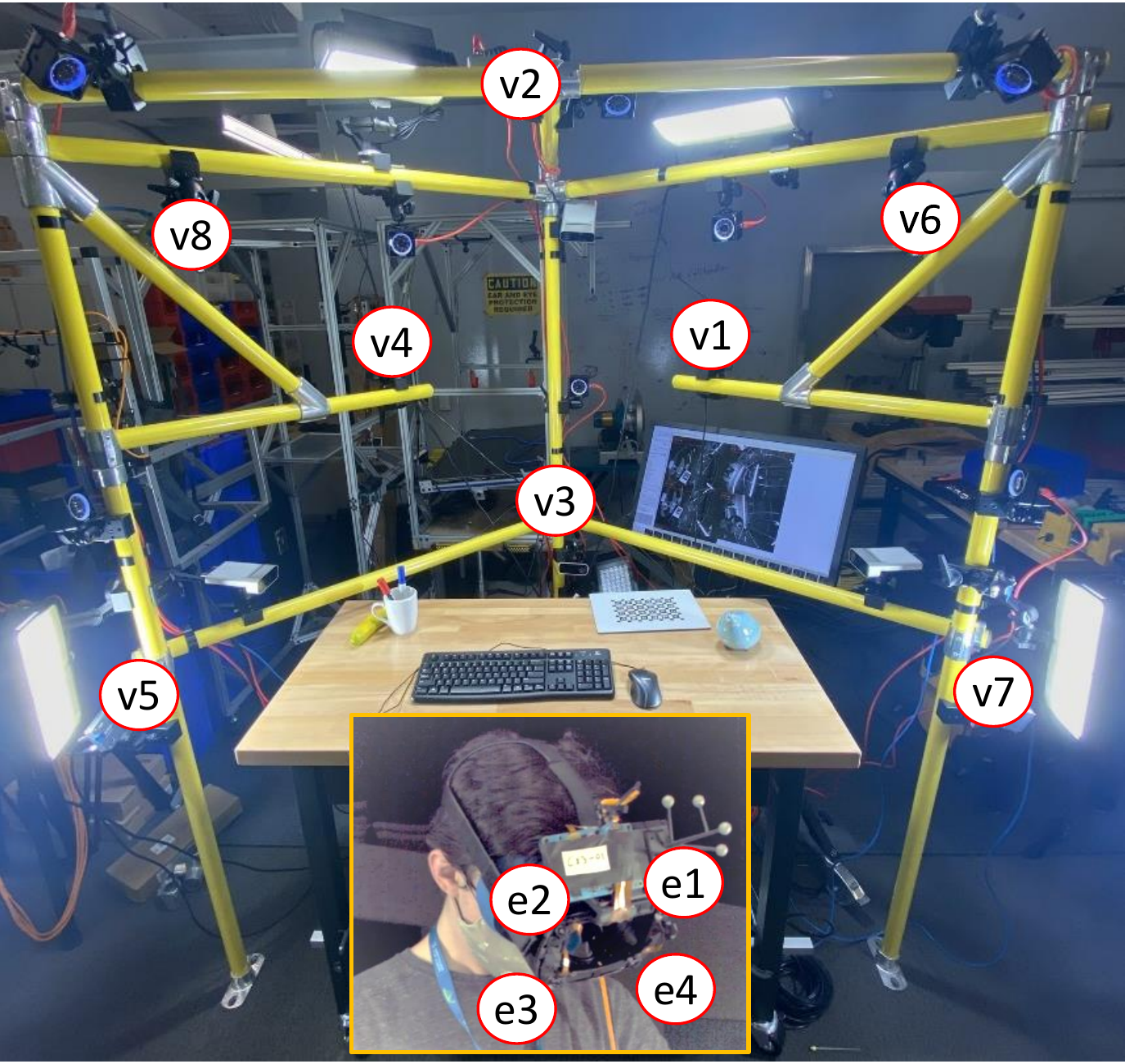} 
\vspace{-2mm}
\caption{Our custom build headset (inset) and multi-camera desk rig, with cameras marked by red circles.} 
\label{fig:cage_desc}
\end{figure}

\begin{figure*}[t]
\centering 
\includegraphics[width=0.45\linewidth]{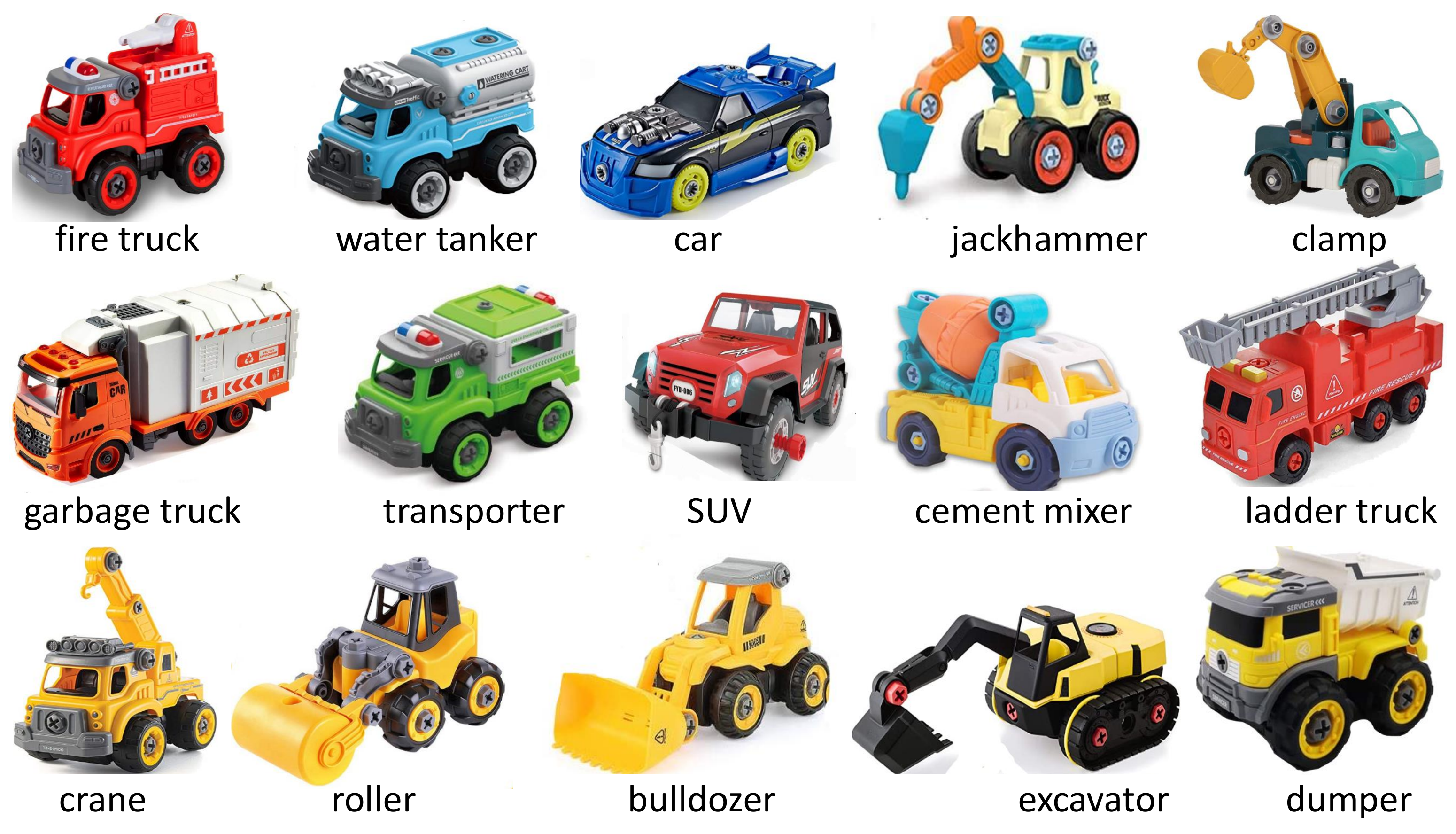} 
\includegraphics[width=0.38\linewidth]{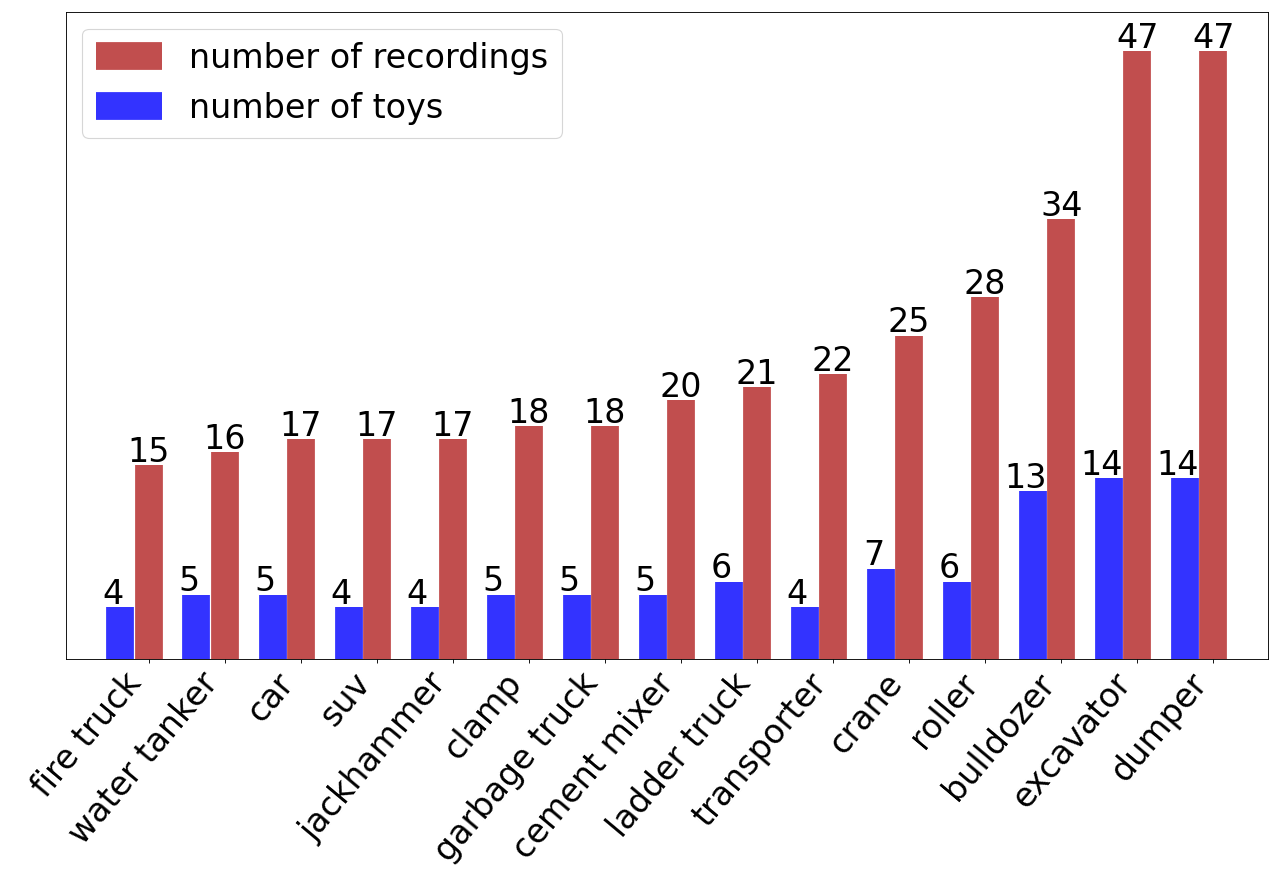}
\vspace{-2mm}
\caption{\textbf{Left:} 15 toy vehicle categories. \textbf{Right:} Distribution of toys and recordings per category. (Best viewed in colour)} 
\vspace{-2mm}
\label{fig:alltoys}
\end{figure*}

\section{Recording and Annotation}
\subsection{Recording rig}
We built a desk rig equipped with eight RGB cameras at $1920\times1080$ resolution and four monochrome cameras at $640\times480$ resolution. The RGB cameras are mounted on a scaffold around the desk with 5 overhead and 3 on the side. The monochrome cameras are placed on the four corners of a custom-built headset worn by the participants and provide multiple egocentric views similar to the Oculus Quest VR headset. Fig.~\ref{fig:cage_desc} shows the recording rig and headset, with cameras circled in red. All cameras are synchronized with SMPTE timecode and geometrically calibrated with a fiducial to sub-pixel accuracies. Participants are recorded standing, though taller participants are asked to sit to ensure their hands and the assembled toy is visible in all camera views. 
 
\subsection{Participants, toys, \& recording protocol}
\noindent\textbf{Participants:} We recruited 53 adults (28 males, 25 females) to disassemble and assemble ``take-apart'' toy vehicles. Each participant was asked to work with six toys in an hour-long recording session, though the final number varies depending on the participant's speed. 

\noindent\textbf{Toys:} The sequences feature 101 unique toys from 15 categories of construction, emergency response, and other vehicles. Each category has variations in colour, size, and style of vehicle; across categories, the vehicles have some shared components \eg construction vehicles feature the same base but different arm attachments. Fig.~\ref{fig:alltoys} shows a sample from each vehicle category and the distribution of toys and recordings per category. 

\noindent\textbf{Protocol:}
We are interested in capturing the \emph{natural} order in which the participants assemble and disassemble the toys, so we placed only an image of the fully assembled toy on the table for reference. We did not provide instructions nor specify a part ordering\footnote{\eg Meccano~\cite{ragusa2021meccano} provides participants with an ordered list of steps.}. This design choice makes the assembly task more challenging but also more realistic, resulting in great variation in action ordering. Preliminary recordings showed that some participants struggled with the assembly task. For time-efficiency, we adjusted the protocol to have participants first disassemble a completed toy before proceeding to ``re''-assemble. 
 
\subsection{Annotations}
\noindent \textbf{Action labels:}
We label two granularities of actions and their start and end times. \emph{\textbf{Fine-grained actions}} are hand-object interactions based on a single verb or movement and an interacting object or toy part. A fine-grained action spans two or three stages: (1) pre-contact when the hand (and tool) starts approaching the object, (2) the interaction, and (3) post-contact when the object is released. Additionally, we merge several co-occuring or sequential fine-grained actions into \emph{\textbf{coarse actions}} related to the attaching or detaching of a vehicle part. For example, the coarse action \emph{``detach bumper''} consists of four fine-grained actions \emph{\{``unscrew bumper with screwdriver'', ``remove screw from bumper'', ``pick up bumper'', ``put down bumper''\}}. The fine-grained actions may overlap with each other as participants often multi-task, \eg, \emph{``put down cabin''} and \emph{``pick up screwdriver''}, while the coarse actions are contiguous. Please see Supplementary for details on annotator training and our custom interface for labelling the actions.
 
\noindent\textbf{3D hand poses:} 
We perform hand tracking from the four monochrome egocentric cameras using a modified version of MegATrack~\cite{han2020megatrack} to estimate 3D hand poses of both hands. First, we fuse features from all views into a shared latent space~\cite{remelli2020lightweight}. Then, we regress the joint angles and global transformation for each hand before obtaining landmarks on the fingertips, joints and palm center via forward kinematics. The tracker is trained end-to-end on the dataset from~\cite{han2020megatrack}. After egocentric tracking, we extract the 3D keypoint locations (21 per hand) in world coordinates as our pose representation (see Fig.~\ref{fig:teaser}). 

\begin{table*}[t]
\centering
\caption{Fine-grained action dataset comparisons.} 
\vspace{-3.5mm}
\resizebox{\textwidth}{!}{
\setlength{\tabcolsep}{1pt}
\begin{tabular}{@{}lp{1pt}rp{1pt}rp{1pt}rp{2pt}|p{2pt}rp{3pt}rp{3pt}rp{2pt}|p{2pt}rp{1pt}rp{1pt}rp{2pt}|p{2pt}rp{3pt}rp{3pt}r@{}}
\toprule
\textbf{Dataset} &&
\textbf{\begin{tabular}[c]{@{}c@{}} total\\hours\end{tabular}} &&
\textbf{\begin{tabular}[c]{@{}c@{}}\#\\videos\end{tabular}} && 
\textbf{\begin{tabular}[c]{@{}c@{}}avg.\\ (min)\end{tabular}} &&&
\textbf{\begin{tabular}[c]{@{}c@{}}\#\\segments\end{tabular}} &&
\textbf{\begin{tabular}[c]{@{}c@{}}avg. \#seg.\\per video\end{tabular}} &&
\textbf{\begin{tabular}[c]{@{}c@{}}avg.\\(sec)\end{tabular}} &&&
\textbf{\begin{tabular}[c]{@{}c@{}}\\verbs\end{tabular}} && 
\textbf{\begin{tabular}[c]{@{}c@{}}\#\\objects\end{tabular}} && 
\textbf{\begin{tabular}[c]{@{}c@{}}\\actions\end{tabular}} &&&
\textbf{\begin{tabular}[c]{@{}c@{}}labelled\\frames\end{tabular}} &&
\textbf{\begin{tabular}[c]{@{}c@{}}overlapping\\segments\end{tabular}} &&
\textbf{\begin{tabular}[c]{@{}c@{}}\#partici-\\pants\end{tabular}} \\ 
\midrule 
Meccano~\cite{ragusa2021meccano} && 6.9 && 20 && 20.7 &&& 8,858 && 442.9 && 2.8 &&& 12 && 21 && 61 &&& 84.9\% && 15.8\% && 20 \\ 
IKEAASM~\cite{ben2021ikea} && 35.0 && 371 && 5.6 &&& 17,577 && 47.3 && 6.0 &&& 12 && 10 && 33 &&& 83.8\% && - && 48 \\ 
EPIC-KITCHENS-100~\cite{Damen2020RESCALING} && 100.0 && 700 && 8.5 &&& 89,977 && 128.5 && 3.1 &&& 97 && 300 && 4,053 &&& 71.6\% && 28.1\% && 37 \\ 
Ego4D~\cite{Ego4D2021} && 120.0 && - && - &&& 77,002 && - && - &&& 74 && 87 &&- &&&- &&- && 406\\
\rowcolor{ao(english)!10}
Assembly101 (ego) && 167.0 && 1,425 && 7.1 &&& 331,310 && 236.7 && 1.7 &&& 24 && 90 && 1,380 &&& 81.4\% && 7.0\% && 53 \\
\rowcolor{ao(english)!10}
Assembly101 && 513.0 && 4,321 && 7.1 &&& 1,013,523 && 236.7 && 1.7 &&& 24 && 90 && 1,380 &&& 81.4\% && 7.0\% && 53 \\ 
\bottomrule
\end{tabular}
}
\label{tab:fine_level_comp}
\end{table*}
 
\begin{table*}[t]
\centering
\caption{Coarse action label dataset comparisons.} 
\vspace{-3.5mm}
\resizebox{\textwidth}{!}{
\setlength{\tabcolsep}{1pt}
\begin{tabular}{@{}lp{8pt}rp{5pt}rp{5pt}cp{5pt}|p{5pt}rp{5pt}cp{5pt}cp{8pt}|p{8pt}rp{5pt}rp{5pt}rp{5pt}|p{5pt}r@{}}
\toprule
\textbf{Dataset} && 
\textbf{\begin{tabular}[c]{@{}c@{}}total\\hours\end{tabular}} && 
\textbf{\begin{tabular}[c]{@{}c@{}}\#\\videos\end{tabular}} && 
\textbf{\begin{tabular}[c]{@{}c@{}}avg. video\\length (min)\end{tabular}} &&&
\textbf{\begin{tabular}[c]{@{}c@{}}\#\\segments\end{tabular}} && 
\textbf{\begin{tabular}[c]{@{}c@{}}avg. \#segments\\per video\end{tabular}} &&
\textbf{\begin{tabular}[c]{@{}c@{}}avg. segments\\length\end{tabular}} &&
\textbf{\begin{tabular}[c]{@{}c@{}}\\verbs\end{tabular}} &&& 
\textbf{\begin{tabular}[c]{@{}c@{}}\#\\objects\end{tabular}} && 
\textbf{\begin{tabular}[c]{@{}c@{}}\\actions\end{tabular}} &&& 
\textbf{\begin{tabular}[c]{@{}c@{}}\#partici-\\pants\end{tabular}}\\ 
\midrule
50Salads~\cite{stein2013combining} && 4.5 && 50 && 6.4 &&& 899 && 18 && 36.8 &&& 6 && 15 && 17 &&& 25 \\ 
Breakfast~\cite{kuehne2014language} && 77.0 && 1,712 && 2.3 &&& 11,300 && 6.6 && 15.1 &&& 14 && 28 && 48 &&& 52 \\ 
\rowcolor{ao(english)!10}
Assembly101 && 513.0 && 4,321 && 7.1 &&& 104,759 && 24 && 16.5 &&& 11 && 61 && 202 &&& 53 \\
\bottomrule
\end{tabular}
}
\label{tab:coarse_level_comp}
\end{table*}

\begin{figure*}[t]
\centering 
\includegraphics[width=0.33\linewidth]{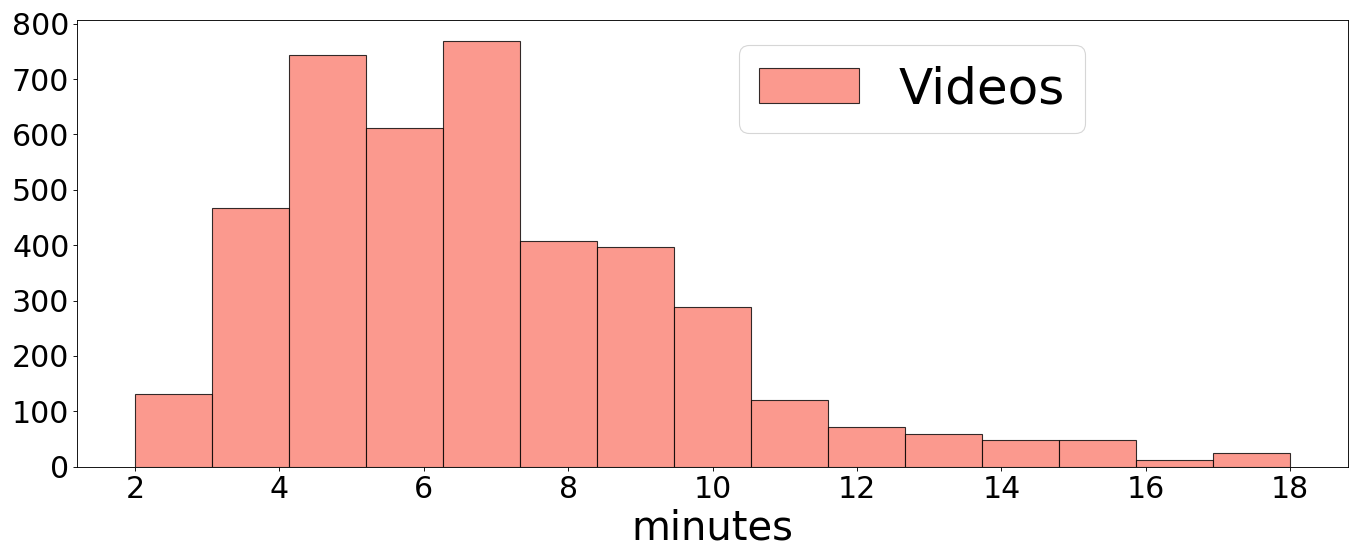} 
\includegraphics[width=0.33\linewidth]{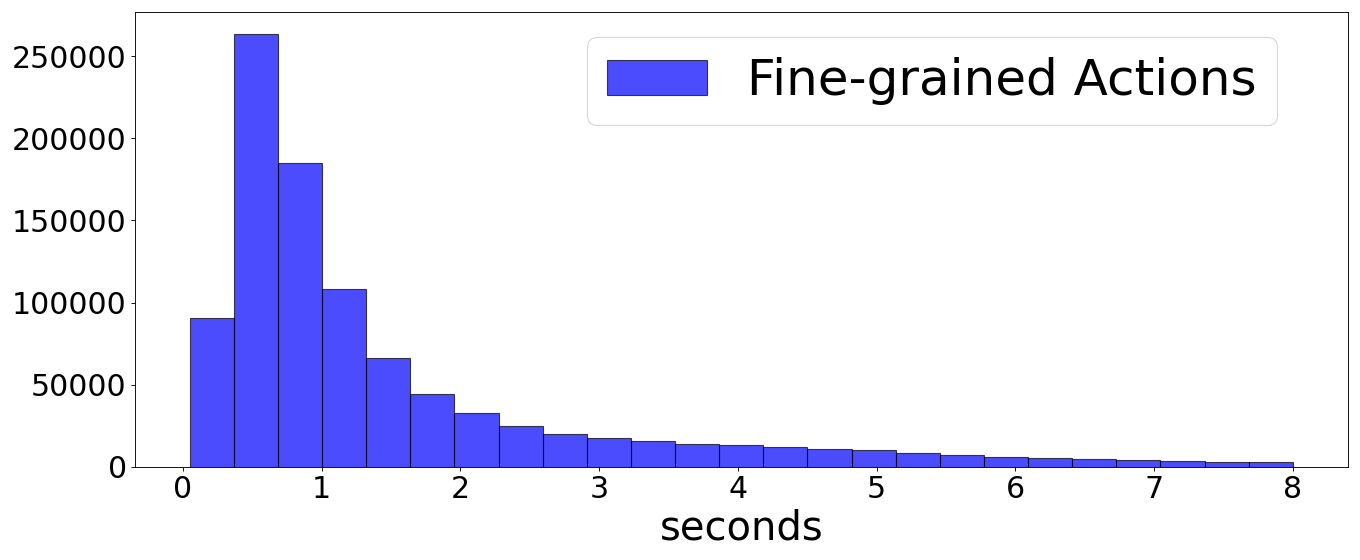} 
\includegraphics[width=0.33\linewidth]{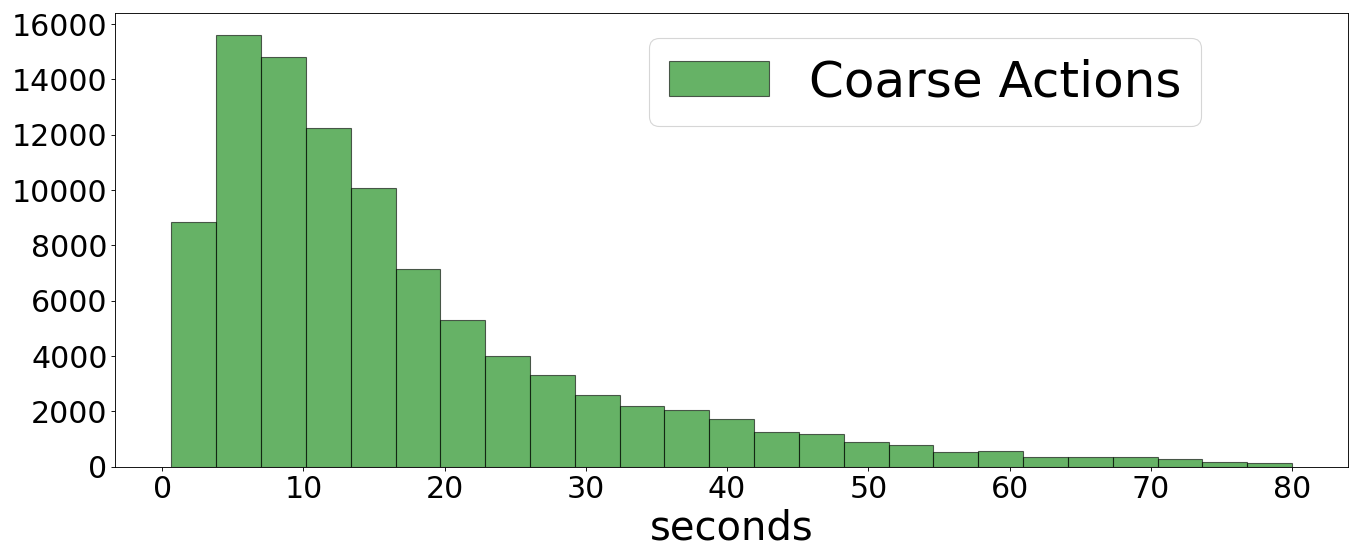}
\vspace{-3mm}
\caption{Distribution of durations: average durations are 7.1 mins, 1.7s and 16.5s for videos, fine-grained and coarse actions, respectively. 
}
\label{fig:segment_video_dur}
\end{figure*}

\section{Dataset Statistics} 
\subsection{Recording statistics}
Our key motivation was to gather a large and diverse procedural activity dataset with varying label granularities. Assembly101 features 362 disassembly-assembly sequences; each sequence is recorded from 12 viewpoints, totalling 4321 videos and 513 hours of footage. The average sequence or video duration is 7.1 $\pm$ 3.4 minutes (Fig.~\ref{fig:segment_video_dur} left). Tables~\ref{tab:fine_level_comp} and \ref{tab:coarse_level_comp} show comparisons with similar recorded datasets. Assembly101 is considerably larger with more than 1M fine-grained and 100K coarse segments, making it the largest procedural activity dataset to date.

\subsection{Fine-grained actions}
From our 15 toy categories, we define 90 objects, \eg, wheel, including 5 tools together with the \emph{``hand''}. Additionally, we specify 24 interaction verbs. The objects and interaction verbs form a total of 1380 fine-grained action labels. Fig.~\ref{fig:segment_video_dur} shows the duration distribution. The average fine-grained action lasts 1.7$\pm$2 seconds. In a single disassembly-assembly sequence, there are an average 236.7$\pm$98.4 fine-grained actions. The entire dataset totals more than 1M fine-grained action instances. The distribution of objects and verbs is provided in the Supplementary. There is a natural long tail, where 30\% of the data accounts for 1238 (89\%) of the fine-grained actions.

\noindent\textbf{Comparison with other datasets:} Table~\ref{tab:fine_level_comp} gives a detailed numerical comparison with other fine-grained action datasets. Assembly101 has 23-44$\times$ more action classes and 56-111$\times$ more action segments than assembly-style datasets IKEA and Meccano. Assembly101's scale is comparable to other large-scale egocentric datasets such as EPIC-KITCHENS and Ego4D. Compared to EPIC, Assembly101's has 1.7$\times$ more egocentric footage and 11$\times$ more action segments. In the labelled footage of Ego4D, the closest subtask of ``forecasting'' features 120 hours of annotated temporal action labels. In comparison, our dataset has 12$\times$ more action segments than Ego4D.

\subsection{Coarse actions}\label{coarse_actions}
Each coarse action is defined by the assembly or disassembly of a vehicle part. There are 202 coarse actions composed of 11 verbs and 61 objects. Each video sequence features an average of 24 coarse actions. The average coarse action comprises 10 fine-grained actions and lasts 16.5 $\pm$ 15.7 seconds (see distribution in Fig.~\ref{fig:segment_video_dur}). We also define the tail classes for the coarse labels where the 30\% of the data accounts for 171 (84\%) of the coarse actions. 

\noindent\textbf{Comparison with other datasets:} 
While coarse actions can also be used for classification, we consider them sequentially and use them for action segmentation. Table~\ref{tab:coarse_level_comp} compares Assembly101 with Breakfast \& 50Salads, two contemporary segmentation benchmarks. We have 2.5$\times$ more videos, 6.7$\times$ more hours of footage, 9.3$\times$ more action segments and 4.2$\times$ more action classes than Breakfast. 
 
\begin{table}[]
\centering
\caption{Temporal dynamics of coarse action segments}
\vspace{-3mm}
\resizebox{\linewidth}{!}{
\setlength{\tabcolsep}{1pt}
\begin{tabular}{@{}lp{20pt}cp{20pt}c@{}}
\toprule
\textbf{Dataset} && \textbf{repetitions} && \textbf{order variations} \\
\midrule 
Breakfast~\cite{kuehne2014language} && 0.11 && 0.15\\ 
50Salads~\cite{stein2013combining} && 0.08 && 0.02\\
\rowcolor{ao(english)!10} Assembly101 && 0.18 && 0.05 \\
\rowcolor{ao(english)!10}
Assembly101 - Assembly && 0.23 && 0.04 \\
\rowcolor{ao(english)!10} Assembly101 - Disassembly && 0.11 && 0.05\\ 
\bottomrule 
\end{tabular}
}
\label{tab:dataset_stats}
\end{table}

\noindent\textbf{Temporal dynamics:}
We define and report two scores in Table~\ref{tab:dataset_stats} to quantify the temporal dynamics. The \textbf{repetition score} is defined as $1-u_i / g_i$ where $u_i$ is the number of unique actions in video $i$, and $g_i$ is the total number of actions and results in a score in the range $[0, 1)$. 0 indicates no repetition, and the closer the score is to 1, the more repetition that occurs in the sequence. Averaged over all video sequences, we have a repetition score of 0.18, with higher repetition (0.23) in assembly than disassembly (0.11). Compared with Breakfast and 50Salad, our dataset includes 1.6$\times$ and 2.3$\times$ more repeated steps, respectively. We compute the \textbf{order variation} as the average edit distance, $e(R,G)$, between every pair of sequences, $(R,G)$, and normalize it with respect to the maximum sequence length of the two, $1-e(R,G)/\text{max}(|R|, |G|)$. This score has a range $[0,1]$; a score of 1 corresponds to no deviations in ordering between pairs. The relatively high scores of Breakfast, 0.15, indicate that actions following a strict ordering, making it less attractive to study temporal sequence dynamics than 50Salads (0.02) and Assembly101 (0.05). Overall, our dataset includes a high frequency of repeated steps and variations in temporal ordering both in assembly and disassembly sequences, which are characteristic of daily procedural activities, and therefore contributes a challenging benchmark for modelling the temporal relations between actions. 

\subsection{Mistake actions}
Even though our participants are adults assembling children's toys, they still make mistakes and then need to make corrections before proceeding. For example, putting on the cabin before attaching the interior (see Fig.~\ref{fig:teaser}), making it impossible to place the interior after, so one must remove the cabin as a corrective action before placing the interior. We annotate the coarse assembly segments with a parallel set of labels \emph{\{``correct'', ``mistake'', ``correction''\}}. 

Mistakes are natural occurrences in many tasks and an opportunity for an AR assistant to provide help. To the best of our knowledge, there are no existing action datasets for recognizing mistakes. Of the 60k coarse actions in assembly, 15.9\% and 6.7\% segments are mistake and corrective segments, respectively. Skill is closely related, but datasets focusing on skill assessment assign a score to short clips of \eg drawing~\cite{doughty2019pros} or suturing~\cite{zia2016automated} instead of determining what and when the mistake occurs. We also annotated the skill level of the participant in our videos from 1 (worst) to 5 (best). Overall, the distribution of skill labels in our sequences is 9\%, 6\%, 13\%, 25\% and 47\% from worst to best. 

\begin{table}[t]
\centering
\caption{Comparisons with other datasets with 3D hand pose.} 
\vspace{-3mm}
\resizebox{\linewidth}{!}{
\setlength{\tabcolsep}{1pt}
\begin{tabular}{@{}lp{9pt}rp{10pt}rp{10pt}rp{10pt}r@{}}
\toprule
\textbf{Dataset} && 
\textbf{total hours} && 
\textbf{\#frames} && 
\textbf{\#segments} && 
\textbf{\#actions} \\ 
\midrule 
FPHA~\cite{garcia2018first} && 1.0 && 0.1M && 1K && 45\\
H2O~\cite{kwon2021h2o} && 5.5 && 0.5M && 1K && 36 \\
\rowcolor{ao(english)!10}
Assembly101 && 513.0&& 111M && 82K && 1456 \\ 
\bottomrule
\end{tabular}
}
\label{tab:hand_pose_comp}
\end{table}

\subsection{3D hand poses}
As Assembly101 features hand-object interactions, 3D hand pose is an important modality, especially since AR/VR systems often provide this information~\cite{han2020megatrack}. Compared with FPHA~\cite{garcia2018first} \& H2O~\cite{kwon2021h2o}, our dataset includes 82$\times$ more segments and 200$\times$ more frames, reported in Table~\ref{tab:hand_pose_comp}. 

\subsection{Training, validation \& test splits}
We use the 60\%, 15\% and 25\% of the videos for creating our training, validation and test splits, respectively, with detailed statistics given in Supplementary. For more robust evaluation, we will withhold the test split ground truths to be used in online submission leaderboards. The validation and test sets are structured to help assess generalization to new toys and actions and the participants' skills. 25 of the 101 toys are shared across training, validation and test splits. There are also toy instances that are not a part of the training set to facilitate zero-shot learning.

\begin{table*}[tb]
\centering
\caption{\textbf{Action recognition} on fine-grained actions evaluated by Top-1 accuracy. \textbf{Action anticipation} on fine-grained actions evaluated by Top-5 Recall.}
\vspace{-3mm}
\resizebox{1\textwidth}{!}{
\setlength{\tabcolsep}{1pt}
\begin{tabular}{@{}ccc|p{1pt}rp{2pt}rp{2pt}rp{2pt}|p{2pt}rp{2pt}rp{2pt}rp{2pt}|p{2pt}rp{2pt}rp{2pt}rp{2pt}|p{2pt}rp{2pt}rp{2pt}rp{2pt}|p{2pt}rp{2pt}rp{2pt}r@{}}
\toprule
& & & 
\multicolumn{7}{c|}{\textbf{Overall}} &
\multicolumn{7}{c|}{\textbf{Head}\cellcolor{ao(english)!10}} & 
\multicolumn{7}{c|}{\textbf{Tail}\cellcolor{ao(english)!10}} & 
\multicolumn{7}{c|}{\textbf{Seen Toys}\cellcolor{pastelred!10}} & 
\multicolumn{6}{c}{\textbf{Unseen Toys}\cellcolor{pastelred!10}} \\ \midrule
Task & Tested on &&& 
verb && object && action &&\cellcolor{ao(english)!10}& 
\cellcolor{ao(english)!10}verb &\cellcolor{ao(english)!10}&\cellcolor{ao(english)!10} object &\cellcolor{ao(english)!10}&\cellcolor{ao(english)!10} action &\cellcolor{ao(english)!10}&\cellcolor{ao(english)!10}&\cellcolor{ao(english)!10}verb &\cellcolor{ao(english)!10}&\cellcolor{ao(english)!10} object &\cellcolor{ao(english)!10}&\cellcolor{ao(english)!10} action &\cellcolor{ao(english)!10}&\cellcolor{pastelred!10}&\cellcolor{pastelred!10}verb &\cellcolor{pastelred!10}&\cellcolor{pastelred!10} object \cellcolor{pastelred!10}&\cellcolor{pastelred!10}&\cellcolor{pastelred!10} action &\cellcolor{pastelred!10}&\cellcolor{pastelred!10}&\cellcolor{pastelred!10}verb &\cellcolor{pastelred!10}&\cellcolor{pastelred!10} object &\cellcolor{pastelred!10}&\cellcolor{pastelred!10} action \\ \midrule
& Fixed &&& 64.0 && 50.4 && 39.2 &&\cellcolor{ao(english)!10}&\cellcolor{ao(english)!10} 69.7 &\cellcolor{ao(english)!10}&\cellcolor{ao(english)!10} 63.3 &\cellcolor{ao(english)!10}&\cellcolor{ao(english)!10} 51.1 &\cellcolor{ao(english)!10}&\cellcolor{ao(english)!10}&\cellcolor{ao(english)!10} 49.7 &\cellcolor{ao(english)!10}&\cellcolor{ao(english)!10} 18.3 &\cellcolor{ao(english)!10}&\cellcolor{ao(english)!10} 9.3 &\cellcolor{ao(english)!10}&\cellcolor{pastelred!10}&\cellcolor{pastelred!10} 63.0 &\cellcolor{pastelred!10}&\cellcolor{pastelred!10} 55.3 &\cellcolor{pastelred!10}&\cellcolor{pastelred!10} 42.0 &\cellcolor{pastelred!10}&\cellcolor{pastelred!10}&\cellcolor{pastelred!10} 64.3 &\cellcolor{pastelred!10}&\cellcolor{pastelred!10} 48.8 &\cellcolor{pastelred!10}&\cellcolor{pastelred!10} 38.3 \\
Recognition & Egocentric &&& 47.0 && 34.3 && 23.0 &&\cellcolor{ao(english)!10}& \cellcolor{ao(english)!10}51.3 &\cellcolor{ao(english)!10}&\cellcolor{ao(english)!10} 44.6 &\cellcolor{ao(english)!10}&\cellcolor{ao(english)!10} 31.0 &\cellcolor{ao(english)!10}&\cellcolor{ao(english)!10}&\cellcolor{ao(english)!10} 36.2 &\cellcolor{ao(english)!10}&\cellcolor{ao(english)!10} 8.6 &\cellcolor{ao(english)!10}&\cellcolor{ao(english)!10} 3.1 &\cellcolor{ao(english)!10}&\cellcolor{pastelred!10}&\cellcolor{pastelred!10} 47.3 &\cellcolor{pastelred!10}&\cellcolor{pastelred!10} 36.0 &\cellcolor{pastelred!10}&\cellcolor{pastelred!10} 23.5 &\cellcolor{pastelred!10}&\cellcolor{pastelred!10}&\cellcolor{pastelred!10} 46.9 &\cellcolor{pastelred!10}&\cellcolor{pastelred!10} 33.8 &\cellcolor{pastelred!10}&\cellcolor{pastelred!10} 22.9 \\
& Fixed \& Ego. &&& 58.5 & &45.2 && 34.0 &&\cellcolor{ao(english)!10}&\cellcolor{ao(english)!10} 63.7 &\cellcolor{ao(english)!10}&\cellcolor{ao(english)!10} 57.2 &\cellcolor{ao(english)!10}&\cellcolor{ao(english)!10} 44.6 &\cellcolor{ao(english)!10}&\cellcolor{ao(english)!10}&\cellcolor{ao(english)!10} 45.3 &\cellcolor{ao(english)!10}&\cellcolor{ao(english)!10} 15.1 &\cellcolor{ao(english)!10}&\cellcolor{ao(english)!10} 7.3 &\cellcolor{ao(english)!10}&\cellcolor{pastelred!10}&\cellcolor{pastelred!10} 57.8 &\cellcolor{pastelred!10}&\cellcolor{pastelred!10} 48.9 &\cellcolor{pastelred!10}&\cellcolor{pastelred!10} 35.9 &\cellcolor{pastelred!10}&\cellcolor{pastelred!10}&\cellcolor{pastelred!10} 58.7 &\cellcolor{pastelred!10}&\cellcolor{pastelred!10} 44.0 &\cellcolor{pastelred!10}&\cellcolor{pastelred!10} 33.3 \\
\bottomrule 
& Fixed &&& 56.6 && 33.3 && 10.4 &&\cellcolor{ao(english)!10}&\cellcolor{ao(english)!10} 60.3 &\cellcolor{ao(english)!10}&\cellcolor{ao(english)!10} 58.1 &\cellcolor{ao(english)!10}&\cellcolor{ao(english)!10} 30.7 &\cellcolor{ao(english)!10}&\cellcolor{ao(english)!10}&\cellcolor{ao(english)!10} 52.8 &\cellcolor{ao(english)!10}&\cellcolor{ao(english)!10} 32.8 &\cellcolor{ao(english)!10}&\cellcolor{ao(english)!10} 6.7 &\cellcolor{ao(english)!10}&\cellcolor{pastelred!10}&\cellcolor{pastelred!10} 55.6 &\cellcolor{pastelred!10}&\cellcolor{pastelred!10} 51.1 &\cellcolor{pastelred!10}&\cellcolor{pastelred!10} 16.9 &\cellcolor{pastelred!10}&\cellcolor{pastelred!10}&\cellcolor{pastelred!10} 56.9 &\cellcolor{pastelred!10}&\cellcolor{pastelred!10} 24.4 &\cellcolor{pastelred!10}&\cellcolor{pastelred!10} 8.2 \\
Anticipation & Egocentric &&& 51.9 && 21.4 && 5.5 &&\cellcolor{ao(english)!10}& \cellcolor{ao(english)!10}54.8 &\cellcolor{ao(english)!10}&\cellcolor{ao(english)!10} 49.6 &\cellcolor{ao(english)!10}&\cellcolor{ao(english)!10} 22.4 &\cellcolor{ao(english)!10}&\cellcolor{ao(english)!10}&\cellcolor{ao(english)!10} 49.2 &\cellcolor{ao(english)!10}&\cellcolor{ao(english)!10} 21.6 &\cellcolor{ao(english)!10}&\cellcolor{ao(english)!10} 2.4 &\cellcolor{ao(english)!10}&\cellcolor{pastelred!10}&\cellcolor{pastelred!10} 51.6 &\cellcolor{pastelred!10}&\cellcolor{pastelred!10} 28.3 &\cellcolor{pastelred!10}&\cellcolor{pastelred!10} 7.9 &\cellcolor{pastelred!10}&\cellcolor{pastelred!10}&\cellcolor{pastelred!10} 51.9 &\cellcolor{pastelred!10}&\cellcolor{pastelred!10} 19.4 &\cellcolor{pastelred!10}&\cellcolor{pastelred!10} 5.3 \\
& Fixed \& Ego. &&& 55.1 && 29.4 && 8.8 &&\cellcolor{ao(english)!10}&\cellcolor{ao(english)!10} 58.5 &\cellcolor{ao(english)!10}&\cellcolor{ao(english)!10} 55.3 &\cellcolor{ao(english)!10}&\cellcolor{ao(english)!10} 28.0 &\cellcolor{ao(english)!10}&\cellcolor{ao(english)!10}&\cellcolor{ao(english)!10} 51.6 &\cellcolor{ao(english)!10}&\cellcolor{ao(english)!10} 29.1 &\cellcolor{ao(english)!10}&\cellcolor{ao(english)!10} 5.3 &\cellcolor{ao(english)!10}&\cellcolor{pastelred!10}&\cellcolor{pastelred!10} 54.3 &\cellcolor{pastelred!10}&\cellcolor{pastelred!10} 43.5 &\cellcolor{pastelred!10}&\cellcolor{pastelred!10} 13.9 &\cellcolor{pastelred!10}&\cellcolor{pastelred!10}&\cellcolor{pastelred!10} 55.3 &\cellcolor{pastelred!10}&\cellcolor{pastelred!10} 22.8 &\cellcolor{pastelred!10}&\cellcolor{pastelred!10} 7.3\\
\bottomrule 
\end{tabular}
}
\label{tab:fineSOA_actionrec}
\end{table*}

\begin{table*}[!thb]
\centering
\caption{Top-1 fine-grained action recognition accuracy for individual views, using TSM networks.}
\vspace{-3mm}
\resizebox{1\textwidth}{!}{
\setlength{\tabcolsep}{1pt}
\begin{tabular}{@{}cp{12pt}|p{12pt}rp{12pt}rp{12pt}rp{12pt}rp{12pt}rp{12pt}rp{12pt}rp{12pt}rp{12pt}rp{12pt}|rp{12pt}rp{12pt}rp{12pt}rp{12pt}rp{12pt}r@{}}
\toprule
\textbf{Trained on} &&& \textbf{v1} && \textbf{v2} && \textbf{v3} && \textbf{v4} && \textbf{v5} && \textbf{v6} && \textbf{v7} && \textbf{v8} && \textbf{all v*} && \textbf{e1} && \textbf{e2} && \textbf{e3} && \textbf{e4} && \textbf{all e*}\\ \midrule
Fixed &&& 43.1 && 40.6 && 40.3 && 43.6 && 27.8 && 40.4 && 33.3 && 37.5 && 38.3 && 1.7 && 1.8 && 2.2 && 3.1 && 2.2 \\
Egocentric &&& 8.1 && 7.5 && 4.8 && 6.0 && 2.9 && 10.8 && 2.6 && 8.5 && 6.4 && 13.2 && 13.2 && 29.2 && 29.3 && 21.2 \\
Fixed \& Ego. &&& 44.1 && 42.6 && 41.1 && 44.8 && 28.0 && 41.5 && 33.4 && 38.2 && 39.2 && 13.9 && 13.1 && 32.7 && 32.7 && 23.0 \\
\bottomrule 
\end{tabular}
}
\label{tab:fine_action_trainings}
\end{table*}

\section{Benchmark Experiments}
We benchmark and present baselines for four action tasks: recognition, anticipation, temporal segmentation and our newly defined mistake recognition. However, as the data is very rich, it is our hope that the extended community will find other uses and tasks for the dataset after its release. Due to limited space, we highlight some key results in this section and defer the architecture, implementation and detailed comparison of results to the Supplementary. 

\subsection{Recognition, anticipation \& segmentation} 
\noindent\textbf{For action recognition} (Table~\ref{tab:fineSOA_actionrec}), we define a classification task on the fine-grained action classes, using pre-trimmed clips based on the annotated start and end times. We train a state-of-the-art video recognition model, TSM~\cite{lin2019tsm}, and two top-performing graph convolutional networks on poses, 2s-AGCN~\cite{shi2019two} and MS-G3D~\cite{liu2020disentangling}. Performance is evaluated by Top-1 accuracies for verb, object and action classes. 

\noindent\textbf{Action anticipation} (Table~\ref{tab:fineSOA_actionrec}), predicts upcoming fine-grained actions $\tau\!=\!1$ second into the future. We train a state-of-the-art model TempAgg~\cite{sener2020temporal}. Performance is evaluated by class-mean Top-5 recall as per~\cite{Damen2020RESCALING}. 
 
\noindent\textbf{Temporal action segmentation} (Table~\ref{tab:SOA_segmentation}) assigns frame-wise action labels to a video sequence. We apply two competing state-of-the-art temporal convolutional networks: MS-TCN++~\cite{li2020ms} and C2F-TCN~\cite{singhania2021coarse}, using frame-wise features extracted from TSM~\cite{lin2019tsm} trained for action recognition on Assembly as input. Performance is evaluated by mean frame-wise accuracy (MoF), segment-wise edit distance (Edit) and F1 scores at overlapping thresholds of 10\%, 25\%, and 50\%, denoted by F1@{10, 25, 50}. 

These three challenges form the basis for understanding actions at various granularities. Compared to the existing datasets, Assembly101 shows great potential for extending video understanding to new challenging natural procedural activities by uniting multi-view recognition, generalization to new tasks, long-tail distributions, different skill levels and sequences with mistakes in one dataset. 
 
\begin{table}[tb]
\centering
\caption{Baselines of \textbf{temporal action segmentation}; unless specified, results are from C2F-TCN. 
}
\vspace{-2mm}
\resizebox{1\linewidth}{!}{
\setlength{\tabcolsep}{1pt}
\begin{tabular}{lp{0pt}cp{2pt}cp{6pt}cp{6pt}cp{6pt}cp{6pt}c}
\toprule
\multicolumn{3}{c}{\textbf{Comparison}} & \multicolumn{6}{c}{\textbf{F1@\{10,25,50\}}} && \textbf{Edit} && \textbf{MoF} \\ \hline
\textbf{SOTA} \\ 
\midrule
MS-TCN++~\cite{li2020ms} && all && 31.6 && 27.8 && 20.6 && 30.7	&& 37.1\\
C2F-TCN~\cite{singhania2021coarse} && all && \textbf{33.3} && \textbf{29.0} && \textbf{21.3} && \textbf{32.4} && \textbf{39.2}\\
\midrule \midrule 
\multicolumn{12}{l}{ \textbf{Fixed vs. Egocentric } } \\ \hline 
Fixed && && 35.5 && 31.2 && 23.2 && 33.9 && 41.3	 \\ 
Egocentric && && 28.7 && 24.4 && 17.5 && 29.2 && 34.8 \\ 
\midrule \midrule 
\multicolumn{12}{l}{ \textbf{Seen vs. Unseen Toys} } \\\hline
Seen && Disassembly && 35.8 && 31.1 && 22.2 && 31.7 && 39.8\\
Unseen && Disassembly && 31.9 && 26.6 && 17.0 && 27.9 && 38.9 \\
\hline
Seen & & Assembly && 33.0 && 28.6 && 22.7 && 30.0 && 42.5 \\ 
Unseen & & Assembly &&	29.9 && 26.2 && 19.8 && 32.0 && 34.8 \\
\bottomrule
\end{tabular}
}
\label{tab:SOA_segmentation}
\vspace{-2mm}
\end{table}

\subsection{Camera viewpoints}
We train the models on the instances from both fixed and egocentric views but report the performance on each view separately in Tables~\ref{tab:fineSOA_actionrec} and~\ref{tab:SOA_segmentation}. Unsurprisingly, fixed viewpoints perform better than egocentric viewpoints, with a difference of 16.2\% in \emph{``Overall''} recognition, 4.9\% recall in \emph{``Overall''} anticipation and 6.5\% MoF in segmentation. These differences highlight the challenging nature of recognizing actions from the egocentric point of view. 

Table~\ref{tab:fine_action_trainings} compares Top-1 action recognition accuracy on the individual camera views. Overhead cameras v4 and v1 have the highest accuracy while side cameras v5 and v7 have the lowest, with a drop of 16\% and 11\% from v4. In egocentric views, the lower headset cameras, e3 and e4 achieve higher accuracies than e1 and e2, which do not fully capture the table. The accuracies of e3 and e4, however, are still more than 10\% lower than that of v4.

Table~\ref{tab:fine_action_trainings} shows that there is a large domain gap if we train the models on only egocentric or fixed view sequences and cross-test rather than training on both sources of data. TSM trained only on fixed views performs significantly worse on egocentric views and vice versa. This indicates a significant mismatch and presents a new challenge for studying the domain gap on paired egocentric and third-person actions. 

\subsection{Head vs. tail classes}
A separate tally in Table~\ref{tab:fineSOA_actionrec} reveals a significant gap of 37\% between head and tail action accuracy for \textbf{recognition}. The drop in tail verbs is much less than objects (18\% vs. 42\% drop). Similarly, the action \textbf{anticipation} performance on head classes is quite high, with a 28\% recall in Table~\ref{tab:fineSOA_actionrec}. It is significantly larger than \emph{``Overall''} action recall by 19.2\%. This large difference could be due to the evaluation metric where the class-mean balances the long-tail distribution as 89\% of action classes are tail classes. Similarly, we evaluate the tail and head class MoF for temporal action segmentation. According to this the MoF of the tail classes is 51.5\% which is much higher than the tail MoF of 7.2\%. The low tail performance scores encourage developing few-shot action recognition methods.

\subsection{Seen vs. unseen, assembly vs. disassembly} 
Assembly101 can be used to study generalization to new assembly tasks through the \emph{``Unseen''} toys. Both Tables~\ref{tab:fineSOA_actionrec} and~\ref{tab:SOA_segmentation} show that \emph{``Seen''} toys score higher than \emph{``Unseen''} ones for action recognition, anticipation and segmentation. For recognition and anticipation, there is little difference in verb scores, but a large gap for objects, as all verbs are shared whereas objects are not (13\% unseen objects). 

We separate the evaluation for assembly vs. disassembly portion of the sequences in Table~\ref{tab:SOA_segmentation} for action segmentation. The MoF and segment scores of the assembly portion is consistently lower than disassembly sequences, likely due to its higher complexity, as the disassembly portions have fewer ordering variations and no mistakes. Overall, the F1 and Edit scores do not show a significant over-segmentation effect compared to disassembly sequences even though the assembly tasks are more complex. 

\begin{table}[t]
\centering
\caption{\textbf{Action recognition} on 3D hand poses.}
\vspace{-3.0mm}
\resizebox{\linewidth}{!}{
\setlength{\tabcolsep}{4.5pt}
\begin{tabular}{@{}llll@{}}
\toprule
\textbf{Method} & \textbf{verb} & \textbf{object} & \textbf{action} \\ \midrule
2s-AGCN~\cite{shi2019two} & 58.1 & 30.9 & 22.2 \\
2s-AGCN~\cite{shi2019two} w/ context & 64.4 & 33.9 & 26.7 \\ 
MS-G3D~\cite{liu2020disentangling} w/ context & \textbf{65.7} & \textbf{36.3} & \textbf{28.7} \\\midrule 
TSM egocentric (fuse 4 views) & 59.0 & 46.5 & 33.8 \\ \midrule\midrule 
Object GT & 28.1& 98.8 &27.2\\ 
MS-G3D~\cite{liu2020disentangling} w/ context + Object GT & \textbf{63.4} & \textbf{98.8} & \textbf{62.0}\\
\bottomrule
\end{tabular}
}
\label{tab:hand_3d_action}
\end{table}

\begin{table}[tb]
\centering
\caption{Frame-wise features are extracted from TSMs pre-trained on various datasets. \textbf{Action recognition} is performed by TempAgg~\cite{sener2020temporal} trained on these features.}
\vspace{-3mm}
\resizebox{1\linewidth}{!}{
\setlength{\tabcolsep}{7.0pt}
\begin{tabular}{@{}lccc@{}}
\toprule 
\textbf{Pre-trained on} & \textbf{verb} & \textbf{object} &
\textbf{action} \\ \midrule
Kinetics-400~\cite{kay2017kinetics} & 28.0 & 19.9 & 9.8 \\
SSv2~\cite{goyal2017something} & 28.7 & 18.8 & 10.2 \\
EPIC-KITCHENS-100~\cite{Damen2020RESCALING} & 44.0 & 25.2 & 17.3 \\
Assembly101 & 65.9 & 50.5 & 40.5 \\ \midrule\midrule 
3D pose - MS-G3D~\cite{liu2020disentangling} w/ context & 65.7 & 36.3 & 28.7 \\ \bottomrule
\end{tabular}
}
\label{tab:ar_domain_gap}
\vspace{-2mm}
\end{table}

\subsection{3D pose-based action recognition}
Another objective for collecting Assembly101 was to investigate action recognition using 3D hand poses. Hand poses are commonly available in AR/VR systems and are significantly more compact representations than video features. Table~\ref{tab:hand_3d_action} compares 3D pose-based to video-based recognition. \emph{``2s-AGCN~\cite{shi2019two}''} classifies trimmed segments bounded by action start and end $[t_s, t_e]$. \emph{``2s-AGCN~\cite{shi2019two} w/ context''} extends each boundary by 0.5 seconds; the extension improves action accuracy significantly. State-of-the-art ``MS-G3D~\cite{liu2020disentangling} w/ context'' achieves the highest action performance of 28.7\%, though this is still 5.1\% lower than the video-based \emph{``TSM egocentric (fuse 4 views)''}, where predictions from the four egocentric views are fused by average voting. Interestingly, the verb accuracy for pose-based recognition is 6.7\% higher than video-based, while its object score is 10.2\% lower than video-based. This is unsurprising as hand poses can easily encode movements but cannot provide much object information. We also add an oracle experiment incorporating ground truth object labels as one-hot encoded frame-level features and train a TempAgg~\cite{sener2020temporal} model on top. As shown in Table ~\ref{tab:hand_3d_action}, \emph{``Object GT''} alone achieves a high object but poor verb accuracy. Fusing it with ``MS-G3D~\cite{liu2020disentangling} w/ context'' results in a significant jump in action accuracy. We leave as future work the joint modeling of 3D objects and hand poses for action recognition.

3D poses have the additional advantage of less sensitivity to domain gaps between different environments. For video-based models, training features from scratch requires considerable amounts of time and data, but using features extracted from pre-trained networks may not always generalize. Table~\ref{tab:ar_domain_gap} compares TempAgg~\cite{sener2020temporal} trained on the features extracted from TSM networks pre-trained on Kinetics-400~\cite{kay2017kinetics}, Something-Something~\cite{goyal2017something}, EPIC-KITCHENS-100~\cite{Damen2020RESCALING} and Assembly101 for view \emph{``v1''}. TSM features pre-trained on EPIC-KITCHENS perform significantly better than the other datasets; though there is still a gap of 23.2\% compared to pre-training on the native Assembly101. This indicates a considerable domain gap between our dataset and the existing action recognition benchmarks. On the other hand, poses are low-dimensional common representations independent of the domain and therefore outperform the scores from the other datasets by a significant margin. 

\subsection{Skill level}
\begin{figure}[t]
\centering 
\includegraphics[width=0.96\linewidth]{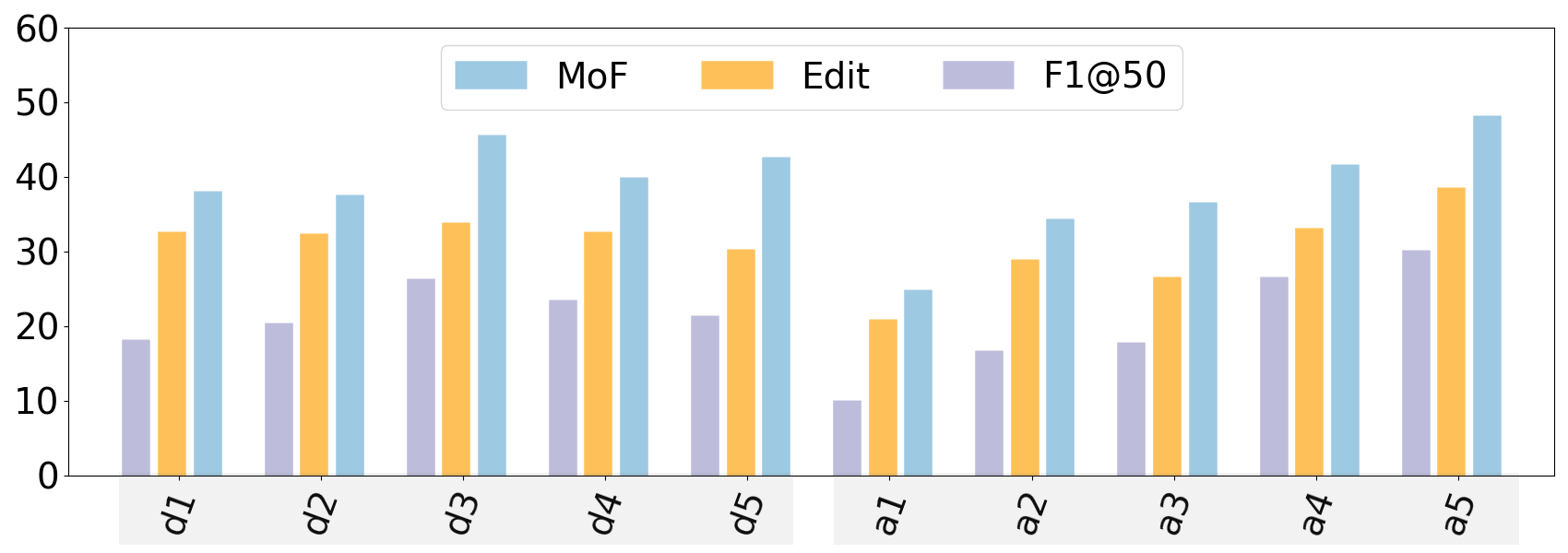} 
\vspace{-3.0mm}
\caption{Influence of skill on segmentation. ``d'' stands for disassembly and ``a'' for assembly. Participants with less skills, ``a1 \& a2'', have lower scores in assembly sequences.}
\label{fig:skillvssegment}
\end{figure}

Fig.~\ref{fig:skillvssegment} compares the segmentation scores for different skill levels from 1 (least skilled) to 5 (most skilled) in both disassembly and assembly sequences, indicated by the prefixes \emph{``d''} and `\emph{``a''}, respectively. Results show that the skill level has little impact on the disassembly sequences. For the least skilled groups \emph{``a1 \& a2''}, however, segmentation scores for assembly sequences are significantly lower than disassembly, likely due to the high ordering variations and mistake segments. 

\subsection{Mistake detection}
Identifying mistakes requires modelling procedural knowledge and retaining long-range sequence information. As input, we provide video sequences represented by frame-wise features from the start of the assembly sequence to the (end of the) current coarse action segment. The task is predicting if the current segment belongs to one of the three classes of \emph{\{``correct'', ``mistake'', ``correction''\}}.
We apply the long-range video model TempAgg~\cite{sener2020temporal} using TSM features and evaluate per-class Top-1 precision and Top-1 recall under two settings: \emph{``Recognition''}, which gets the entire coarse segment and \emph{``Early prediction''}, which gets half of the segment. Due to the imbalanced class distribution, we penalize the models more for misclassifying ``mistake'' and ``correction'' classes. As an oracle baseline, we use the ground truth coarse action labels, \emph{``GT coarse''} as input. 

\noindent\textbf{Baseline results:}
Table~\ref{tab:mistake_detection} shows the challenge in detecting mistakes - even using the ground truth coarse action labels as input, the recall for mistakes and corrections is only around 62.7\% and 84.9\% respectively. With TSM input features, the recall is currently only around 46.6\% and 29.6\% once the segment of interest ends. Early prediction results in a further 11.6\% and 3.2\% drop. 

\begin{table}[tb]
\centering
\caption{Mistake detection results.}
\vspace{-3mm}
\resizebox{1\linewidth}{!}{
\setlength{\tabcolsep}{3.0pt}
\begin{tabular}{@{}llcccc@{}}
\toprule
\textbf{} & \textbf{} & \multicolumn{2}{c}{\textbf{Mistake}} & \multicolumn{2}{c}{\textbf{Correction}} \\ \midrule
\textbf{Task} & \textbf{Features} & precision & recall & precision & recall \\ \midrule
Recognition & GT coarse & 48.6 & 62.7 & 65.6 & 84.9 \\
 & TSM & 30.8 & 46.6 & 30.8 & 29.6 \\ \midrule
Early prediction & TSM & 29.3 & 35.0 & 26.5 & 26.4 \\ \bottomrule
\end{tabular}
}
\label{tab:mistake_detection}
\end{table}
 
\section{Conclusion}
In this paper, we presented Assembly101, the largest procedural activity dataset to date. Our dataset includes synchronized egocentric and static viewpoints for cross-view domain analysis, multi-granular action segments and mistake labels to study goal-oriented sequence learning and 3D hand poses to advance 3D hand-object interaction recognition. We defined four challenges, action recognition, action anticipation, temporal action segmentation and mistake detection, to evaluate a wide range of aspects of assembly tasks, including generalization to new toys, cross-view transfer, long-tailed distributions, skill level and pose vs. appearance. Existing methods show promising results but are still far from tackling these challenges with high precision, as observed in the oracle experiments, leaving room for future explorations. 

Assembly101 can be used for many different applications. In this paper, we proposed several directions such as training the next generation of smart assistants to recognize what a user is doing, predict subsequent steps as they watch an assembly task, check for non-compliant steps and give alerts or offer help. We hope that the community will find other applications and tasks for our dataset after its release.\\

\noindent\textbf{Acknowledgements:}
\small{
This research/project is supported by the National Research Foundation, Singapore under its AI Singapore Programme (AISG Award No: AISG2-RP-2020-016). Any opinions, findings and conclusions or recommendations expressed in this material are those of the author(s) and do not reflect the views of National Research Foundation, Singapore.}

{\small
\bibliographystyle{ieee_fullname}
\bibliography{egbib}
}

\end{document}


\title{Supplementary \\ Assembly101: A Large-Scale Multi-View Video Dataset \\ 
for Understanding Procedural Activities} 

\author{Fadime Sener$^{\dagger}$ \;\; Dibyadip Chatterjee$^{\ddagger}$ \;\; Daniel Shelepov$^{\dagger}$ \;\; Kun He$^{\dagger}$ \;\; Dipika Singhania$^{\ddagger}$ \\ Robert Wang$^{\dagger}$ \;\; Angela Yao$^{\ddagger}$\vspace{10pt}\\ 
		$^\dagger$ Meta Reality Labs Research \;\; 
		$^\ddagger$ National University of Singapore\\ 
  {\tt\small \{famesener,dsh,kunhe,rywang\}@fb.com} \;\;
 {\tt\small \{dibyadip,dipika16,ayao\}@comp.nus.edu.sg}\\ 
		\textcolor{blue}{\url{https://assembly-101.github.io/}\vspace{-6pt}}
}

\maketitle

\begin{strip}
\centering 
\includegraphics[width=0.93\linewidth]{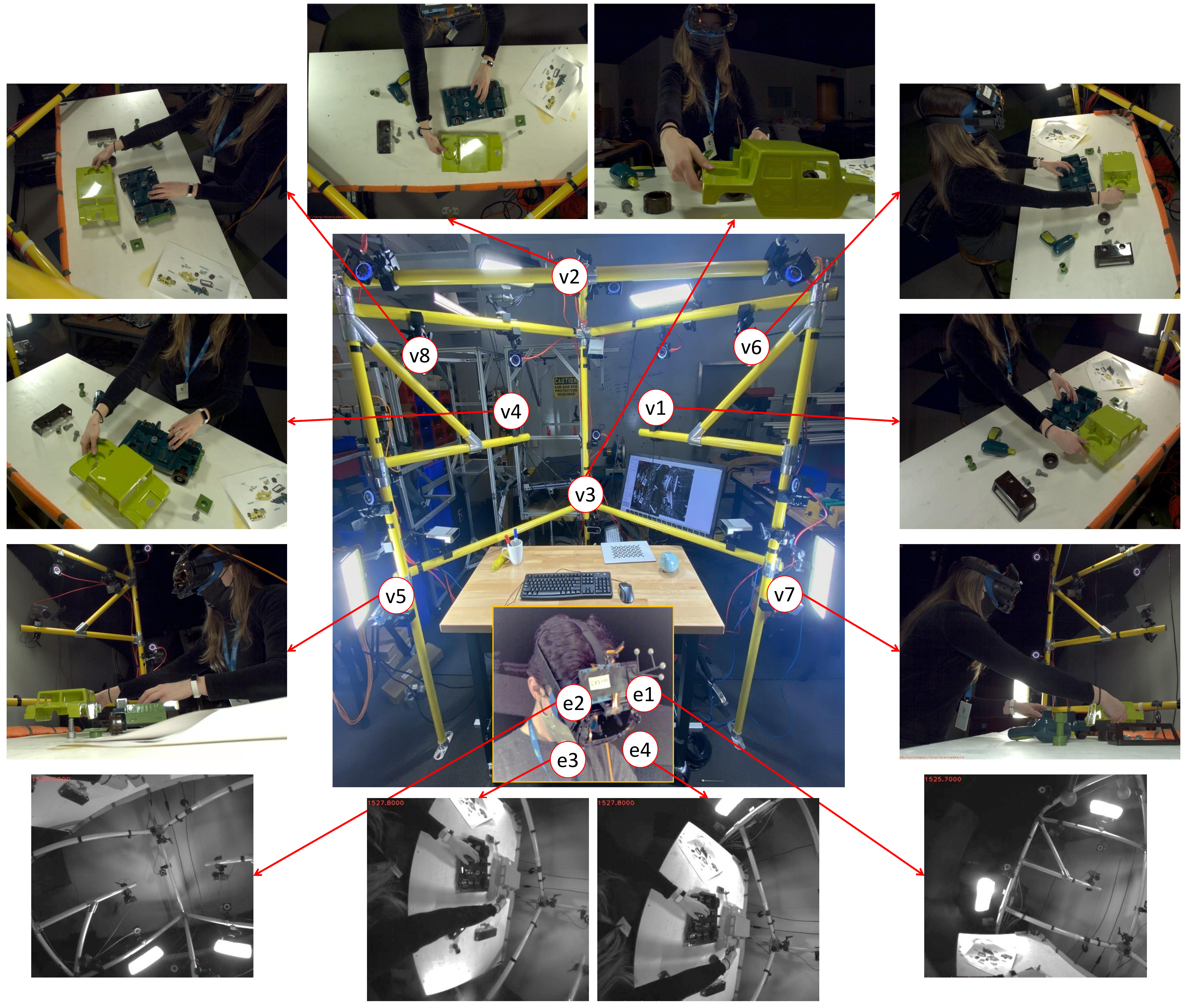} 
\vspace{-2mm}
\captionof{figure}{Our desk-based rig and sample frames from eight RGB and four monochrome cameras.}
\label{fig:sup_allviews} 
\end{strip}

\begin{strip}
\centering 
\includegraphics[width=0.95\linewidth]{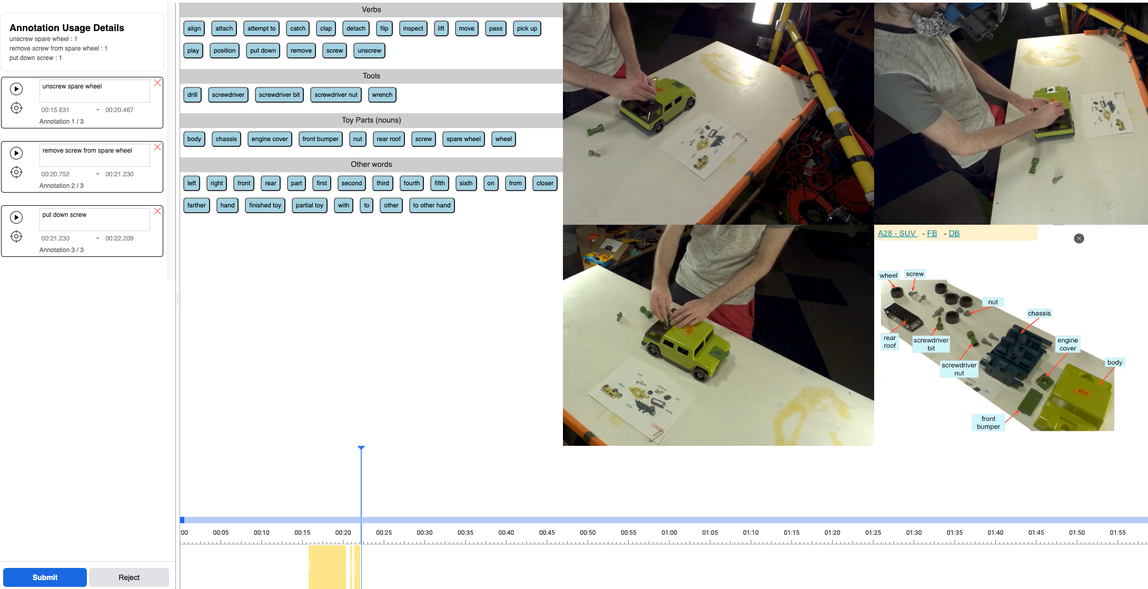} 
\vspace{-1mm}
\captionof{figure}{Our annotation tool interface. \textbf{Right panel}: input video composed of three static camera views and a diagram of the objects. \textbf{Middle panel}: pre-defined lists of verbs, tools and objects for labelling segments. The annotators also have the flexibility for free-form entry. \textbf{Bottom bar}: this annotation bar shows the temporal boundaries of the actions, i.e., the start and end of each action. \textbf{Left panel}: list of temporally annotated actions.} 
\vspace{-2mm}
\label{fig:suu_tool}
\end{strip}

\begin{figure*}[t]
\centering 
\includegraphics[width=0.99\linewidth]{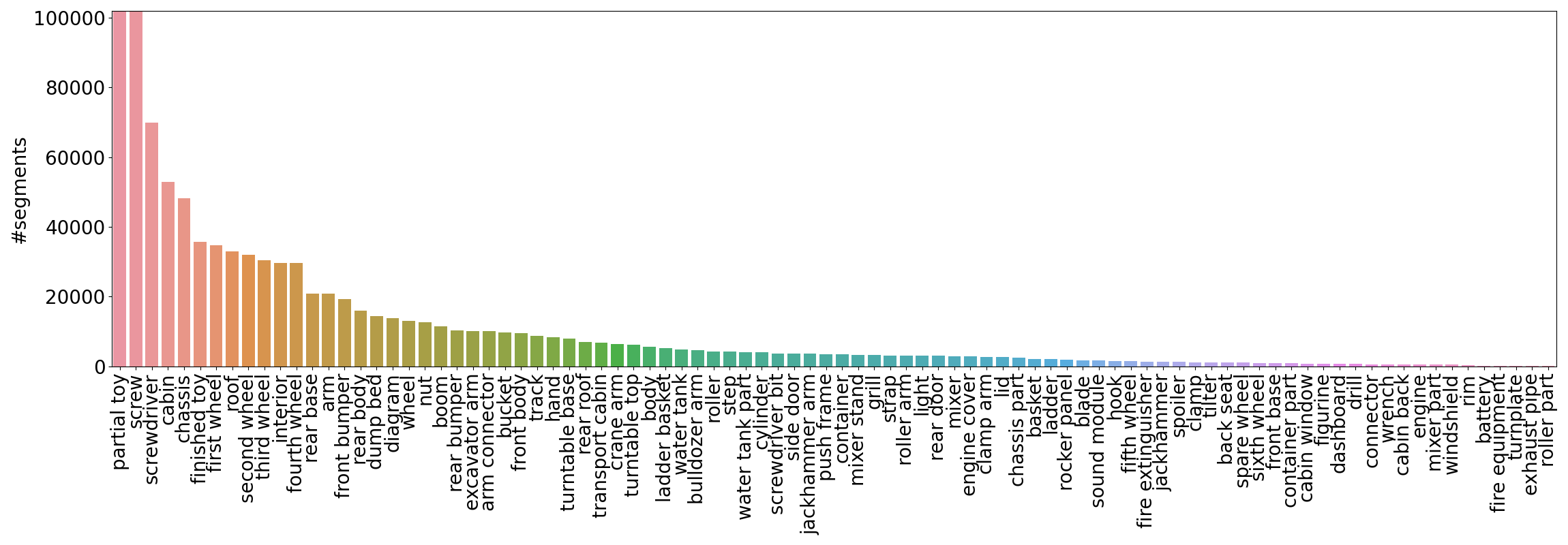} 
\includegraphics[width=0.97\linewidth]{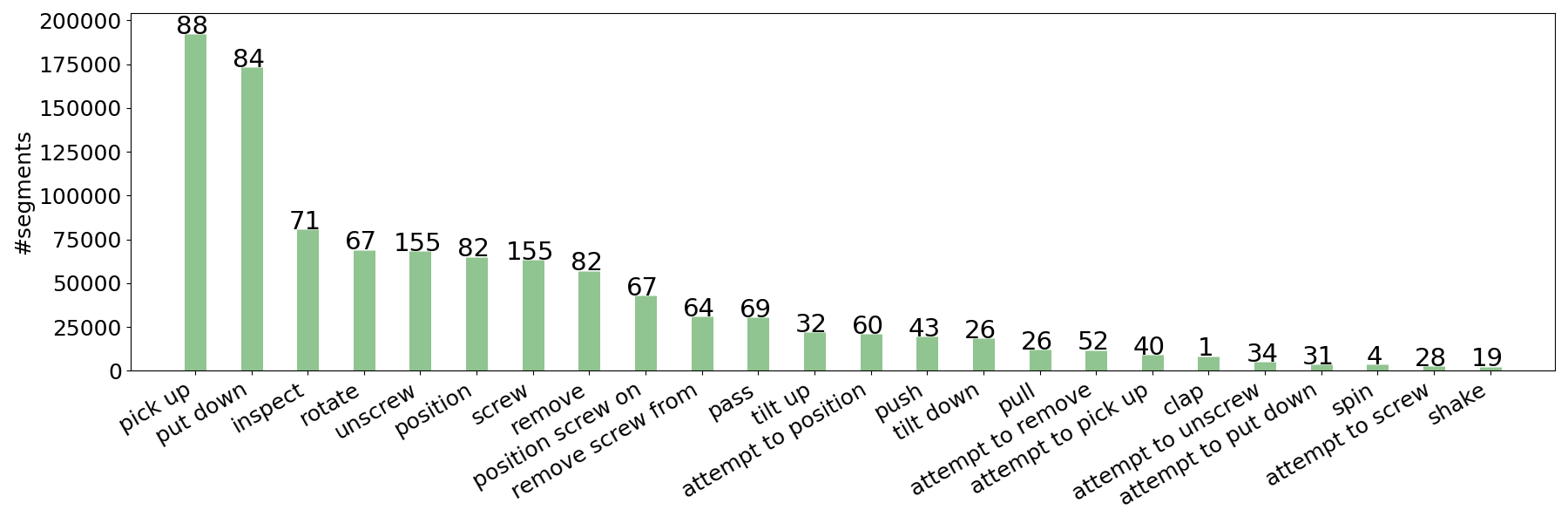} 
\vspace{-2mm}
\caption{We define 90 objects (\textbf{upper}) and specify 24 verbs (\textbf{bottom}), forming a total of 1380 fine-grained action labels. The verb distribution also shows the number of actions containing that verb on top of each bar.}
\label{fig:fine_dists_verb_noun}
\end{figure*}

This supplementary further details recording settings, annotations and experiments. Section~\ref{sup_recording} provides an overview of annotator training, our recordings and custom interface. Section~\ref{supp_stats} provides the distributions of our labels, detailed train/validation/test statistics, and an extensive set of comparisons with other related datasets. Section~\ref{supp_details} presents the architecture and implementation details of our baselines. Finally, in Section~\ref{supp_results}, we present more results and comparisons on our baselines. 

\section{Recording and Annotation}\label{sup_recording}
\subsection{Recording rig}
We built a dedicated desk-based rig to capture the sequences in this dataset. Each sequence is recorded with eight RGB cameras at 1920 × 1080 resolution and four monochrome cameras at 640 × 480 resolution. Fig.~\ref{fig:sup_allviews} shows individual camera views and sample frames. 
 
\subsection{Participants}
We recruited 53 adult participants (28 males, 25 females) to record approximately one-hour sessions over the course of 18 consecutive days. Participants were recruited considering the guidelines and restrictions of COVID-19, including wearing face masks. We obtained informed consent of camera wearers for the digital capture of participants, which means digitally capturing participants’ faces and bodies. All video footage and collected annotations are available for the research community.

\subsection{Annotations}
\vspace{-2mm}
\noindent\paragraph{Annotation interface:} 
We developed a custom interface (see Fig.~\ref{fig:suu_tool}) for annotators to temporally locate the start- and end-frames of fine-grained action segments. Each action segment is tagged with predefined verbs, tools and objects, though annotators also have the flexibility for free-form entry. To promote precise annotations, we displayed three static camera views to ensure that the actions are visible at least from one view without self-occlusion from the working hand. Additionally, we provided diagrams for the annotators with labelled objects of all 101 toys to ensure correct naming and terminology. 
While working with the toys, the participants were requested to simultaneously describe out loud their actions with named tools and objects, \eg \emph{``I am flipping the truck over and putting the right front wheel on the truck''}. To assist with the description, the completed toy in the reference diagrams are labelled with part names.

\noindent\paragraph{Annotator training:} To ensure high-quality labels, we trained annotators over the course of four days. During this time, the annotators were introduced to our interface and the labelling task under the authors' guidance. After training, annotators who were slow or made many mistakes were not selected to continue. Following the training, the labelling was completed by 21 annotators over 213 hours of work.

\section{Dataset Statistics \& Splits}~\label{supp_stats}
\vspace{-6mm}
\subsection{Fine-grained actions}
From our 15 toy categories, we define 90 unique objects. Additionally, we define 24 verbs. Six of the 24 verbs are used to describe ``attempts'', \ie the participants adjust or change their minds during assembly. For example, the ``pick up chassis'' action is composed of three stages of {\it reaching for the chassis}, {\it grasping it}, and {\it lifting it up}. Our annotators were provided with the stages of each verb. When users do not complete all stages in a segment, \eg approach and/or grasp the chassis but do not lift it, we asked our annotators to place “attempt to” in front of the action. The objects and verbs combined form a total of 1380 fine-grained action labels as not every possible combination is observed. We present the distribution of our verbs and objects in Fig.~\ref{fig:fine_dists_verb_noun}. 

To highlight the scale of our dataset, we compare Assembly101 to other video datasets for action recognition in Table~\ref{tab:supp_fine_level_dataset}. Our dataset is the largest in number of segments and the richest in terms of multi-view recordings both from third-person and egocentric views.

\paragraph{Balancing the head:}
The objects, verbs, and fine-grained action labels each naturally form a long-tailed distribution~\cite{simoncelli2001natural}. When reviewing Meccano and IKEA, we observe that a handful of head-classes dominate the action distribution (60\% of actions belong to 3 head classes in IKEA, and 30\% in Meccano). To mitigate similar effects, we made two labelling design choices concerning the wheels, screws and tools, as they are the most commonly occurring object parts. The adjustments spread the head-tail distribution (the top 3 classes account for only 13\% of the action segments) and add semantic richness to the dataset: 

\begin{itemize}
\item \textbf{Enumerating the wheels:}, \ie \emph{``position first wheel''} vs. the generic \emph{``position wheel''} action. Enumeration also extends the range of temporal dependencies in a sequence, as algorithms must keep track of how many wheels have been attached or removed. 
 
\item \textbf{Fine-grained tool and screw verbs:} Due to the nature of the assembly task, tools and screws appear very frequently. To spread the head classes that result from treating tools and screws as simple objects or parts, we introduced dedicated verbs, \eg ``screw [object] with drill", ``position screw on [object]'' and ``remove screw from [object]''. Coupling these verbs with other objects conveys more information than ``screw chassis'' or ``position screw". 
\end{itemize} 
 
\begin{table*}[t]
\centering
\caption{Comparison with other video datasets for action recognition on fine-grained actions.} 
\vspace{-2mm}
\resizebox{\textwidth}{!}{
\setlength{\tabcolsep}{9.0pt}
\begin{tabular}{@{}lrrrr|rrrrr@{}}
\toprule
\textbf{Dataset} & 
\textbf{\begin{tabular}[c]{@{}c@{}} total\\hours\end{tabular}} &
\textbf{\begin{tabular}[c]{@{}c@{}}\#\\videos\end{tabular}} & 
\textbf{\begin{tabular}[c]{@{}c@{}}\#\\segments\end{tabular}} & 
\textbf{\begin{tabular}[c]{@{}c@{}}\#\\actions\end{tabular}} & 
\textbf{\begin{tabular}[c]{@{}c@{}}\\recorded\end{tabular}} &
\textbf{\begin{tabular}[c]{@{}c@{}}multi-\\view\end{tabular}} & 
\textbf{\begin{tabular}[c]{@{}c@{}}\\ egocentric\end{tabular}} &
\textbf{\begin{tabular}[c]{@{}c@{}}\#pose\\ annotation\end{tabular}}&
\textbf{\begin{tabular}[c]{@{}c@{}}\\year\end{tabular}} \\ 
\midrule 
MPII~\cite{rohrbach2012database} & 8.3 & 44 & 5,609 & 64 & \cmark & \xmark & \xmark & \cmark & 2012\\
ActivityNet~\cite{caba2015activitynet} & 648.0 & 27,811 & 23,064 & 200 & \xmark & \xmark & \xmark & \xmark & 2015 \\
Charades~\cite{sigurdsson2016hollywood} & 81.1 & 9,848 & 67,000 & 157 & \cmark & \xmark & \xmark & \xmark & 2016 \\
THUMOS~\cite{idrees2017thumos} & 30.0 & 5,613 & 6,310& 101 & \xmark & \xmark & \xmark & \xmark & 2017\\
Charades-EGO~\cite{sigurdsson2018actor} & 68.8 & 2,751 & 30,516 & 157 & \cmark & \cmark & \cmark & \xmark & 2018 \\
EPIC-100~\cite{Damen2020RESCALING} & 100.0 & 700 & 89,977 & 4053 & \cmark & \xmark & \cmark & \xmark & 2020 \\
H2O~\cite{kwon2021h2o} & 5.5 & 186 & 934 & 11 & \cmark & \cmark & \cmark & \cmark & 2021 \\
Meccano~\cite{ragusa2021meccano} & 6.9 & 20 & 8,858 & 61 & \cmark & \xmark & \cmark & \xmark & 2021 \\
IKEAASM~\cite{ben2021ikea} & 35.0 & 371 & 17,577 & 33 & \cmark & \cmark & \xmark & \cmark & 2021\\
Ego4D~\cite{Ego4D2021} & 120.0 &- &77,002 & - & \cmark & \xmark & \cmark & \xmark & 2021 \\\midrule
\textbf{Assembly101} & 513.0& 4,321 & 1,013,523 & 1380 & \cmark & \cmark & \cmark & \cmark & 2021 \\ \bottomrule
\end{tabular}}
\label{tab:supp_fine_level_dataset}
\end{table*}

\begin{table*}[t]
\centering
\caption{Coarse action label dataset comparisons.} 
\vspace{-2mm}
\resizebox{\textwidth}{!}{
\setlength{\tabcolsep}{5pt}
\begin{tabular}{@{}lrrrr|rrrrr@{}}
\toprule
\textbf{Dataset} & 
\textbf{hours} & 
\textbf{\#videos} & 
\textbf{\#segments} &
\textbf{\#actions} & 
\textbf{\#recorded} & 
\textbf{\#multi-view} & 
\textbf{\#egocentric} & 
\textbf{\#cooking} & 
\textbf{\#year} \\ 
\midrule
GTEA~\cite{fathi2011learning} & 0.4 & 28 & 500 & 71 & \cmark & \xmark & \cmark & \cmark & 2011 \\ 
50Salads~\cite{stein2013combining} & 4.5 & 50 & 899 & 17 & \cmark & \xmark & \xmark & \cmark & 2013\\ 
Breakfast~\cite{kuehne2014language} & 77.0 & 1,712 & 11,300 & 48 & \cmark & \cmark & \xmark & \cmark & 2014\\ 
YouTube Instructional~\cite{alayrac2016unsupervised} & 7.0 & 150 & 1,260 & 47 & \xmark & \xmark & \xmark & \xmark & 2016 \\ 
COIN~\cite{tang2019coin} & 476.0 & 11,800 & 46,000 & 778 & \xmark & \xmark & \xmark & \xmark &2019 \\ 
CrossTask~\cite{zhukov2019cross} & 374.0 & 4,700 & 34,000 & 107 & \xmark & \xmark & \xmark & \cmark & 2019\\
YouCookII~\cite{zhou2018towards} & 176.0 & 2,000 & 15,400 & - & \xmark & \xmark & \xmark & \cmark & 2018 \\ \midrule
\textbf{Assembly101} & 513.0& 4,321 & 104,759 & 202 & \cmark & \cmark & \cmark & \xmark & 2021 \\ \bottomrule
\end{tabular}}
\label{tab:supp_coarse_level_comp}
\end{table*}

\begin{table}[t]
\centering
\caption{Comparisons with other datasets with 3D hand pose.} 
\vspace{-2mm}
\resizebox{\linewidth}{!}{
\setlength{\tabcolsep}{3.0pt}
\begin{tabular}{@{}lrrrr@{}}
\toprule
\textbf{Dataset} & 
\textbf{Hours} & 
\textbf{\#frames} & 
\textbf{\#segments} & 
\textbf{\#actions} \\ 
\midrule 
NTU RGB+D 60~\cite{shahroudy2016ntu} & - & 4.0M & 56K & 60 \\
NTU RGB+D 120~\cite{liu2019ntu} & - & 8.0M & 114K & 120 \\
FPHA~\cite{garcia2018first} & 1.0h & 0.1M & 1K & 45\\
H2O~\cite{kwon2021h2o} & 5.5h & 0.5M & 1K & 36 \\\midrule
\textbf{Assembly101} & 513.0h & 110.0M & 86K & 1380 \\ 
\bottomrule
\end{tabular}}
\label{tab:supp_fine_level_comp}
\end{table}

\begin{table}[!htb]
\centering
\caption{The distribution of \emph{\{``correct'', ``mistake'', ``correction''\}} segments on the coarse segments of the assembly sequences.} 
\vspace{-2mm}
\resizebox{\linewidth}{!}{
\setlength{\tabcolsep}{15pt}
\begin{tabular}{@{}crrr@{}}
\toprule
& \textbf{\#correct} & \textbf{\#mistake} & \textbf{\#correction} \\ \midrule
Test & 12,337 & 3,144 & 1,268 \\
Validation & 8,984 & 1,624 & 640 \\
Train & 25,718 & 4,941 & 2,226\\ \midrule
Overall & 47,039 & 9,709 & 4,134 \\ 
\bottomrule
\end{tabular}
}
\label{tab:supp_mistake_correction}
\end{table}

\begin{table*}[t]
\centering
\caption{Statistics of Assembly101 and its Train/Validation/Test splits.}
\vspace{-2mm}
\resizebox{\textwidth}{!}{
\setlength{\tabcolsep}{5pt}
\begin{tabular}{@{}lrr|rr|rrrr|rrrr@{}}
\toprule
\textbf{Split} & 
\textbf{Hours} & 
\textbf{\#videos} & 
\textbf{\begin{tabular}[c]{@{}c@{}}\#unseen \\toys\end{tabular}} &
\textbf{\begin{tabular}[c]{@{}c@{}}\#shared \\toys\end{tabular}} &
\textbf{\begin{tabular}[c]{@{}c@{}}\#fine\\segments\end{tabular}} &
\textbf{\begin{tabular}[c]{@{}c@{}}\#fine\\verbs\end{tabular}} &
\textbf{\begin{tabular}[c]{@{}c@{}}\#fine\\objects\end{tabular}} &
\textbf{\begin{tabular}[c]{@{}c@{}}\#fine\\actions\end{tabular}} & 
\textbf{\begin{tabular}[c]{@{}c@{}}\#coarse\\segments\end{tabular}} &
\textbf{\begin{tabular}[c]{@{}c@{}}\#coarse\\verbs\end{tabular}} &
\textbf{\begin{tabular}[c]{@{}c@{}}\#coarse\\objects\end{tabular}} &
\textbf{\begin{tabular}[c]{@{}c@{}}\#coarse\\actions\end{tabular}} \\ 
\midrule
Train & 287.6 & 2526 & 40 & 26 & 566,855 & 24 & 85 & 1244 & 57,657 & 11 & 59 & 195 \\ 
Validation & 96.6 & 740 & 16 & 18 & 186,788 & 24 & 81 & 1018 & 19,008 & 10 & 56 & 164 \\ 
Test & 128.8 & 1055 & 20 & 20 & 259,880 & 24 & 79 & 1045 & 28,094 & 11 & 55 & 172 \\ 
\midrule
Overall & 513.0 & 4321 & 76 & 25 & 1,013,523 & 24 & 90 & 1380 & 104,759 & 11 & 61 & 202 \\ 
\bottomrule
\end{tabular}
}
\label{tab:supp_oursplits}
\end{table*}

\subsection{Coarse actions}
Each coarse action is defined by the assembly or disassembly of a vehicle part. There are 202 coarse actions composed of 11 verbs and 61 objects. Each video sequence features an average of 24 coarse actions. There is an average of 10 fine-grained actions per coarse action segment. The average number of coarse actions is 14 in each assembly sequence and 10 in each disassembly sequence. Table~\ref{tab:supp_coarse_level_comp} compares Assembly101 with other video datasets with coarse labels. Our dataset is the largest in video hours and number of segments, and the only non-cooking recorded dataset. 

\subsection{3D hand poses} 
Action recognition from 3D hand poses is much less explored compared to the full human body. The only existing datasets~\cite{garcia2018first,kwon2021h2o} that focus on hand-object action recognition with 3D hand pose annotations are small-scale and/or include only a single hand~\cite{garcia2018first}. We present our comparisons in Table~\ref{tab:supp_fine_level_comp}. Compared with FPHA\cite{garcia2018first} and H2O~\cite{kwon2021h2o}, our dataset includes 82$\times$ more action segments and 200$\times$ more frames. 
We also compare the scale of our dataset with NTU RGB+D 60~\cite{shahroudy2016ntu} and NTU RGB+D 120~\cite{liu2019ntu}, which are the largest full-body pose dataset. Our dataset contains 6-12$\times$ more action classes and 27-13$\times$ more frames. Additionally, NTU RGB+D 60 and NTU RGB+D 120 are composed of short trimmed clips of actions while our segments are related to each other with sequence dynamics, which allows for studying the importance of temporal context for action recognition.
 
\subsection{Training, validation \& test splits}
We use a 60/15/25 split of recordings for dividing our dataset into training, validation and test splits, with detailed statistics presented in Table~\ref{tab:supp_oursplits}. We present the distribution of the mistake action in Table~\ref{tab:supp_mistake_correction}. 

For evaluation purposes, we will hold out the ground truth annotations of the test split. These will be used for online challenge leaderboards to track future progress on our target tasks. Our dataset is designed to assess the generalizability to new toys, actions and the participants' skills. We thus structured our validation and test sets to examine models under varying conditions.

\noindent\textbf{Seen/Unseen vehicles/toys:} Of the 101 toys, only 25 toys are shared across all the three splits. We designed the splits to ensure that there are unseen toys in the training to facilitate zero-shot learning. There are 20 and 16 unseen toy instances in the validation and test splits, respectively. 

\noindent\textbf{Head vs. tail classes:} 
The distribution of our objects and verbs can be seen in Figure~\ref{fig:fine_dists_verb_noun}. There is a large number of common manipulation verbs such as ``pick up'' and ``put down'', which naturally depicts a long tail distribution. The object and action distribution follow the same general trend. We define the tail classes as the set of action classes whose instances account for 30\% of the training data. This amounts to 1238 (89\%) tail action classes. We used Epic-Kitchens as a reference when forming our tail classes, where 87\% of the action classes are in the tail. Similarly, we define the tail classes of the coarse labels as the set of coarse action classes whose instances account for 30\% of the training data. This amounts 171 (84\%) tail action classes. 
 
\begin{figure}[!htb]
\centering 
\includegraphics[width=1\linewidth]{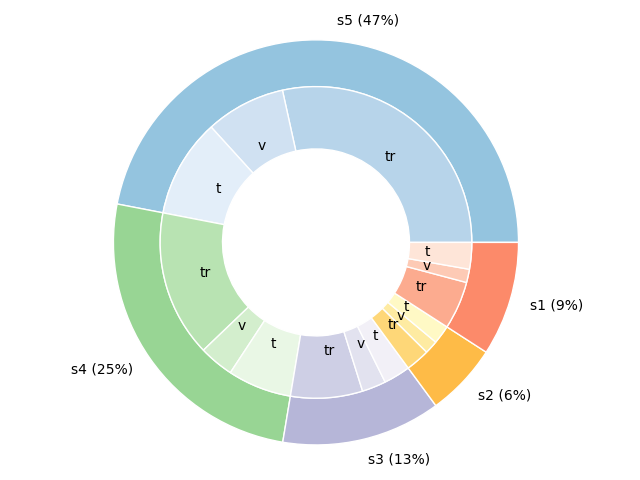} 
\vspace{-8mm}
\caption{The distribution of skill level of the participants from 1 (the worst) to 5 (the best). Overall, 9\% of the sequences are from the participants with the worst skill level and 47\% is from the best. `tr', `v' and `t' stand for the training, validation and test splits. }
\label{fig:supp_skill_level}
\end{figure}

\noindent\textbf{Skill level} assessment is a critical task in many areas including sports~\cite{parmar2017learning}, robot learning~\cite{stadie2017third}, surgery~\cite{vedula2016task} and assembly line~\cite{pedersen2016robot}. Which participant has the highest assembly skills? How are the participants progressing with more assembly tasks? What are the common mistakes made by participants? Answering these questions involves determining how well the assembly was carried out. Thus, we annotated the skill levels of the participant in each video from 1 (the worst) to 5 (the best). Skill level criteria is based on the participant's assembly speed and number of mistakes, with coarse thresholds. Overall, the distribution of skill labels in our sequences is 9\%, 6\%, 13\%, 25\% and 47\% from the worst to the best (see Fig.~\ref{fig:supp_skill_level}).
 
\section{Implementation Details}~\label{supp_details}
We define four action challenges: recognition, anticipation, temporal segmentation, and mistake recognition. 
 
\subsection{Action recognition} 
\subsubsection{Appearance-based action recognition} 
Top-performing video-based action recognition models~\cite{carreira2017quo,feichtenhofer2019slowfast} are typically extensions of state-of-the-art image-based architectures~\cite{he2016deep}. Some works extend convolution and pooling to the time dimension~\cite{carreira2017quo,feichtenhofer2019slowfast}; others perform channel shifting~\cite{lin2019tsm,fanbuch2020rubiks} to capture temporal relationships while maintaining the complexity of a 2D CNN. We adopted a state-of-the-art model, TSM~\cite{lin2019tsm}, as the baseline for this task. \\ 
 
\noindent\textbf{Implementation details:}
We use two versions of the standard TSM architecture with a ResNet-50~\cite{he2016deep} backbone --- one with a single classifier head for predicting the actions and another with two classifier heads for predicting the objects and verbs separately. Both models are trained using stochastic gradient descent (SGD) with a momentum of 0.9, weight decay of 0.0005, and dropout of 0.5 for 50 epochs with a batch size of 64. The learning rate is initialized as 0.001 and decayed by a factor of 10 at epochs 20 and 40. The best-performing model is selected via early-stopping over the validation set. Sampling and augmentation during training and inference for TSM is done following~\cite{Damen2020RESCALING}.

\subsubsection{Pose-based action recognition:}
State-of-the-art methods for recognizing skeleton-based actions are based on deep architectures such as CNNs~\cite{liu2019skepxels}, transformers~\cite{plizzari2021spatial} and graph convolutional networks (GCN)~\cite{yan2018spatial,liu2020disentangling}. We use two state-of-the-art GCN-based methods for our experiment, 2s-AGCN~\cite{shi2019two} and MS-G3D~\cite{liu2020disentangling}. \\

\noindent\textbf{Implementation details:} 
We use the publicly available PyTorch~\cite{paszke2019pytorch} code for 2s-AGCN and MS-G3D. All hand pose sequences are padded to T = 200 frames by replaying the action segments. If there is one hand missing, we pad the second hand with 0. No data augmentation is used. 

We trained 2s-AGCN~\cite{shi2019two} using SGD with Nesterov momentum of 0.9 and a learning rate of 0.1 with a batch size of 32 for 30 epochs. The weight decay is set to 0.0001. For MS-G3D~\cite{liu2020disentangling}, we used SGD with a momentum of 0.9 and a learning rate of 0.05. We set the batch size to 16 and the weight decay to 0.0005. The model is trained for 50 epochs. 
 
\subsection{Action anticipation} 
In our experiments, the anticipation task is defined as predicting the upcoming fine-level actions \emph{1 second} before they start. We adopted TempAgg~\cite{sener2020temporal} as baseline for this task. Similar to previous works~\cite{Damen2020RESCALING,furnari2018leveraging}, we report class-mean Top-5 recall as it accounts for uncertainty in future predictions.\\

\begin{table*}[tb]
\centering
\caption{\textbf{Action recognition} on fine-grained actions evaluated by Top-5 accuracy.} 
\resizebox{1\textwidth}{!}{
\setlength{\tabcolsep}{1pt}
\begin{tabular}{@{}ccc|p{1pt}rp{2pt}rp{2pt}rp{2pt}|p{2pt}rp{2pt}rp{2pt}rp{2pt}|p{2pt}rp{2pt}rp{2pt}rp{2pt}|p{2pt}rp{2pt}rp{2pt}rp{2pt}|p{2pt}rp{2pt}rp{2pt}r@{}}
\toprule
& & & 
\multicolumn{7}{c|}{\textbf{Overall}} &
\multicolumn{7}{c|}{\textbf{Head}\cellcolor{ao(english)!10}} & 
\multicolumn{7}{c|}{\textbf{Tail}\cellcolor{ao(english)!10}} & 
\multicolumn{7}{c|}{\textbf{Seen Toys}\cellcolor{pastelred!10}} & 
\multicolumn{6}{c}{\textbf{Unseen Toys}\cellcolor{pastelred!10}} \\ \midrule
Task & Tested on &&& 
verb && object && action &&\cellcolor{ao(english)!10}& 
\cellcolor{ao(english)!10}verb &\cellcolor{ao(english)!10}&\cellcolor{ao(english)!10} object &\cellcolor{ao(english)!10}&\cellcolor{ao(english)!10} action &\cellcolor{ao(english)!10}&\cellcolor{ao(english)!10}&\cellcolor{ao(english)!10}verb &\cellcolor{ao(english)!10}&\cellcolor{ao(english)!10} object &\cellcolor{ao(english)!10}&\cellcolor{ao(english)!10} action &\cellcolor{ao(english)!10}&\cellcolor{pastelred!10}&\cellcolor{pastelred!10}verb &\cellcolor{pastelred!10}&\cellcolor{pastelred!10} object \cellcolor{pastelred!10}&\cellcolor{pastelred!10}&\cellcolor{pastelred!10} action &\cellcolor{pastelred!10}&\cellcolor{pastelred!10}&\cellcolor{pastelred!10}verb &\cellcolor{pastelred!10}&\cellcolor{pastelred!10} object &\cellcolor{pastelred!10}&\cellcolor{pastelred!10} action \\ \midrule
& Fixed &&& 91.2 && 77.0 && 63.3 &&\cellcolor{ao(english)!10}&\cellcolor{ao(english)!10} 93.8 &\cellcolor{ao(english)!10}&\cellcolor{ao(english)!10} 89.6 &\cellcolor{ao(english)!10}&\cellcolor{ao(english)!10} 78.0 &\cellcolor{ao(english)!10}&\cellcolor{ao(english)!10}&\cellcolor{ao(english)!10} 84.9 &\cellcolor{ao(english)!10}&\cellcolor{ao(english)!10} 45.8 &\cellcolor{ao(english)!10}&\cellcolor{ao(english)!10} 26.4 &\cellcolor{ao(english)!10}&\cellcolor{pastelred!10}&\cellcolor{pastelred!10} 90.8 &\cellcolor{pastelred!10}&\cellcolor{pastelred!10} 84.8 &\cellcolor{pastelred!10}&\cellcolor{pastelred!10} 68.6 &\cellcolor{pastelred!10}&\cellcolor{pastelred!10}&\cellcolor{pastelred!10} 91.4 &\cellcolor{pastelred!10}&\cellcolor{pastelred!10} 74.6 &\cellcolor{pastelred!10}&\cellcolor{pastelred!10} 61.6 \\
Recognition & Egocentric &&& 82.7 && 64.3 && 44.3 &&\cellcolor{ao(english)!10}& \cellcolor{ao(english)!10} 86.0 &\cellcolor{ao(english)!10}&\cellcolor{ao(english)!10} 79.1 &\cellcolor{ao(english)!10}&\cellcolor{ao(english)!10} 57.8 &\cellcolor{ao(english)!10}&\cellcolor{ao(english)!10}&\cellcolor{ao(english)!10} 74.6 &\cellcolor{ao(english)!10}&\cellcolor{ao(english)!10} 27.4 &\cellcolor{ao(english)!10}&\cellcolor{ao(english)!10} 10.8 &\cellcolor{ao(english)!10}&\cellcolor{pastelred!10}&\cellcolor{pastelred!10} 83.5 &\cellcolor{pastelred!10}&\cellcolor{pastelred!10} 67.8 &\cellcolor{pastelred!10}&\cellcolor{pastelred!10} 46.3 &\cellcolor{pastelred!10}&\cellcolor{pastelred!10}&\cellcolor{pastelred!10} 82.5 &\cellcolor{pastelred!10}&\cellcolor{pastelred!10} 63.2 &\cellcolor{pastelred!10}&\cellcolor{pastelred!10} 43.7 \\
& Fixed \& Ego. &&& 88.5 && 72.9 && 57.1 &&\cellcolor{ao(english)!10}&\cellcolor{ao(english)!10} 91.2 &\cellcolor{ao(english)!10}&\cellcolor{ao(english)!10} 86.2 &\cellcolor{ao(english)!10}&\cellcolor{ao(english)!10} 71.4 &\cellcolor{ao(english)!10}&\cellcolor{ao(english)!10}&\cellcolor{ao(english)!10} 81.6 &\cellcolor{ao(english)!10}&\cellcolor{ao(english)!10} 39.8 &\cellcolor{ao(english)!10}&\cellcolor{ao(english)!10} 21.3 &\cellcolor{ao(english)!10}&\cellcolor{pastelred!10}&\cellcolor{pastelred!10} 88.4 &\cellcolor{pastelred!10}&\cellcolor{pastelred!10} 79.2 &\cellcolor{pastelred!10}&\cellcolor{pastelred!10} 61.2 &\cellcolor{pastelred!10}&\cellcolor{pastelred!10}&\cellcolor{pastelred!10} 88.5 &\cellcolor{pastelred!10}&\cellcolor{pastelred!10} 70.9 &\cellcolor{pastelred!10}&\cellcolor{pastelred!10} 55.8 \\
\bottomrule 
\end{tabular}
}
\label{tab:supp_fineSOA_actionrec}
\end{table*}

\begin{table*}[tb]
\centering
\caption{\textbf{Action recognition \& anticipation} performance on fine-level actions (evaluate by Top-1 acc. and Top-5 recall respectively) using TSM and TempAgg respectively. ``Fusion'' corresponds to average-pooling the scores from multiple views.}
\resizebox{1\textwidth}{!}{
\setlength{\tabcolsep}{5pt}
\begin{tabular}{@{}cc|ccc|ccc|ccc|rrr|rrr@{}}
\toprule
 & & \multicolumn{3}{c|}{\textbf{Overall}} & \multicolumn{3}{c|}{\textbf{Head}} & \multicolumn{3}{c|}{\textbf{Tail}} & \multicolumn{3}{l|}{\textbf{Seen Toys}} & \multicolumn{3}{l}{\textbf{Unseen Toys}} \\ \midrule
& & verb & object & act. & verb & object & act. & verb & object & act. & \multicolumn{1}{c}{verb} & \multicolumn{1}{c}{object} & \multicolumn{1}{c|}{act.} & \multicolumn{1}{c}{verb} & \multicolumn{1}{c}{object} & \multicolumn{1}{c}{act.} \\ \midrule
Recognition & Overall & 58.5 & 45.2 & 34.0 & 63.7 & 57.2 & 44.6 & 45.3 & 15.1 & 7.3 & 57.8 & 48.9 & 35.9 & 58.7 & 44.0 & 33.3 \\ 
& Fusion & 71.6 & 59.0 & 48.0 & 77.4 & 74.4 & 63.2 & 57.0 & 20.9 & 10.4 & 71.0 & 64.7 & 51.2 & 71.8 & 57.2 & 46.9 \\ \midrule
Anticipation & Overall & 55.1 & 29.4 & 8.8 & 58.5 & 55.3 & 28.0 & 51.6 & 29.1 & 5.3 & 54.3 & 43.5 & 13.9 & 55.3 & 22.8 & 7.3\\ 
& Fusion & 59.2 & 31.3 & 9.1 & 62.6 & 62.3 & 34.8 & 55.5 & 30.3 & 4.5 & 58.3 & 48.3 & 15.7 & 59.4 & 23.4 & 7.8 \\
\bottomrule 
\end{tabular}
}
\label{tab:supp_fineSOA_anticipation}
\end{table*}

\noindent\textbf{Implementation details:}
We use the TempAgg with three classification heads that predicts objects, verbs and actions separately. Since TempAgg operates on frame features, we use the 2-D backbone of the TSM fine-tuned on our dataset to extract the 2048-D frame features. The {\it spanning past} snippet features are computed over a period of 6 seconds before the start of the action and aggregated at 3 temporal scales $K = \{5,3,2\}$. The {\it recent past} snippet features are computed over a period of $\{1.6, 1.2, 0.8, 0.4\}$ before the start of the action and aggregated over a single temporal scale $K_R = 2$.
The model is trained using an Adam~\cite{kingma2014adam} optimizer for 15 epochs with a batch size of 32. A dropout factor of 0.3 is used. The learning rate initialised as 0.0001 and decayed by a factor of 10 after the 10\textsuperscript{th} epoch. 

\subsection{Temporal action segmentation}
For temporal action segmentation, we apply two competing state-of-the-art temporal convolutional networks: MS-TCN++~\cite{li2020ms}, which maintains a fixed temporal resolution in its feed-forward structure with successively larger kernel dilation, and C2F-TCN~\cite{singhania2021coarse}, a U-net-style shrink-then-expand encoder-decoder architecture. For C2F-TCN, we use implicit ensembling of decoder layers and the feature augmentation strategy detailed in the paper. Performance is evaluated by mean frame-wise accuracy (MoF). Since longer actions dominate this score and it does not penalize over-segmentation errors explicitly, we also report segment-wise edit distance (Edit) and F1 scores at overlapping thresholds of 10\%, 25\%, and 50\%, denoted as by F1@{10, 25, 50}.

\noindent\textbf{Implementation details:} 
For both C2F-TCN~\cite{singhania2021coarse} and MS-TCN++~\cite{li2020ms}, we use an Adam~\cite{kingma2014adam} optimizer with a batch size of 20 for a maximum of 200 epochs while using early-stopping to select the model that best fits the validation data. Loss functions used for both models are frame-wise cross entropy loss weighted with 1 and mean-square error loss~\cite{li2020ms} weighted with 0.17. For MS-TCN++, we use a learning of 0.0005 and a weight decay of 0. For C2F-TCN, we use a learning rate of 0.001 and weight decay of 0.0001. The base window for feature augmentation sampling is set to be 20 and all layers of decoder are included in ensembling.

\subsection{Mistake detection} 
We introduce the new problem of mistake detection in assembly videos. We adopted TempAgg~\cite{sener2020temporal} as the baseline for this task, which captures long-range relationships that span an order of several minutes successfully. \\

\noindent\textbf{Implementation details:}
We modified the TempAgg model to capture even longer-range relationships. More precisely, the {\it spanning past} snippet features are computed over a period of 60 seconds around the action segment, i.e., $[s-60, e+60]$, aggregated at 3 temporal scales $K = \{5,3,2\}$, where $s$ and $e$ are the start and end timestamps of the action in seconds. The {\it recent past} snippet features are computed over a period of $\{3.0, 2.0, 1.0, 0.0\}$ around the action segment and aggregated over a single temporal scale $K_R = 5$. The training scheme remains similar to anticipation, i.e., it is trained on 2048-D TSM features using an Adam~\cite{kingma2014adam} optimizer for 15 epochs with a batch size of 32 and a dropout of 0.3 on a single GPU. The learning rate initialised as 0.0001 decayed by a factor of 10 after the 10\textsuperscript{th} epoch. Due to the imbalanced class distribution, we used a weighted cross-entropy loss to penalize the model more for misclassifying ``mistake'' and ``correction'' classes.

\section{Results}\label{supp_results}
\subsection{Action recognition \& anticipation}
In Table~\ref{tab:supp_fineSOA_actionrec}, we provide Top-5 accuracy for action recognition. We compare our \emph{``Overall''} performance with results obtained by fusing scores from multiple views on recognition and anticipation in Table~\ref{tab:supp_fineSOA_anticipation}. The fusion increases the performance of recognition significantly, while the improvement is smaller for anticipation. 

\begin{figure}[t]
\centering 
\includegraphics[width=0.99\linewidth]{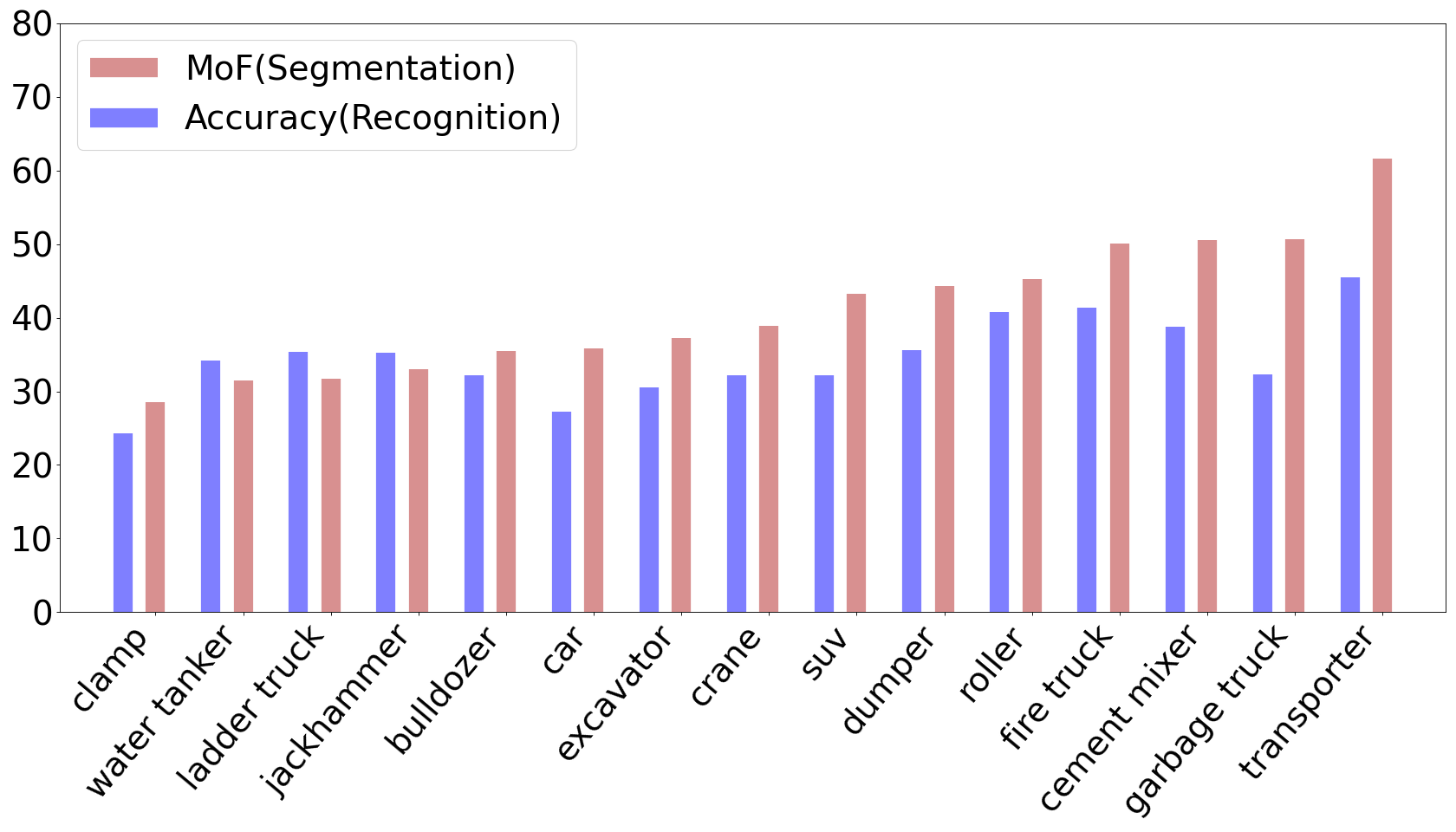} 
\caption{Action recognition accuracy and segmentation MoF over toy categories.}
\label{fig:toy_cats}
\end{figure}

\begin{figure}[t]
\centering 
\includegraphics[width=1\linewidth]{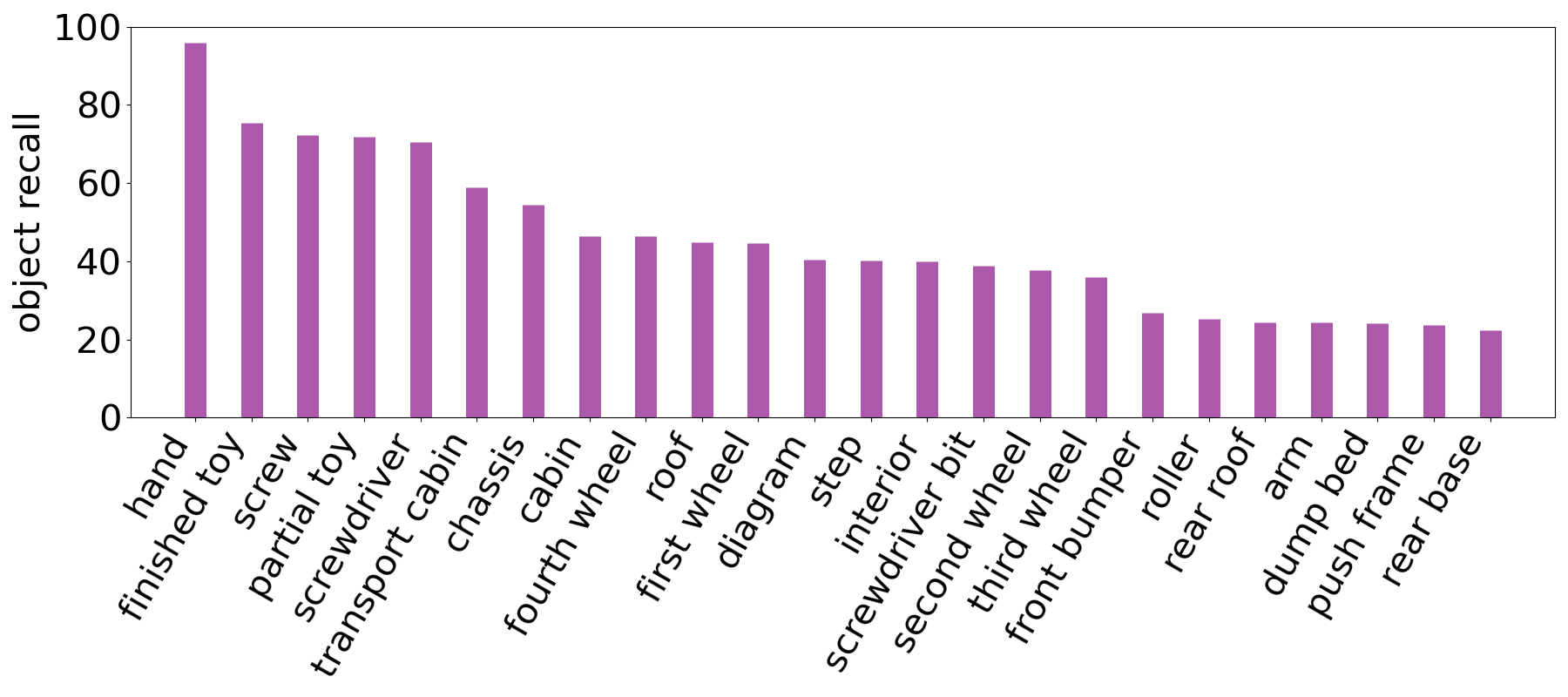} 
\includegraphics[width=1\linewidth]{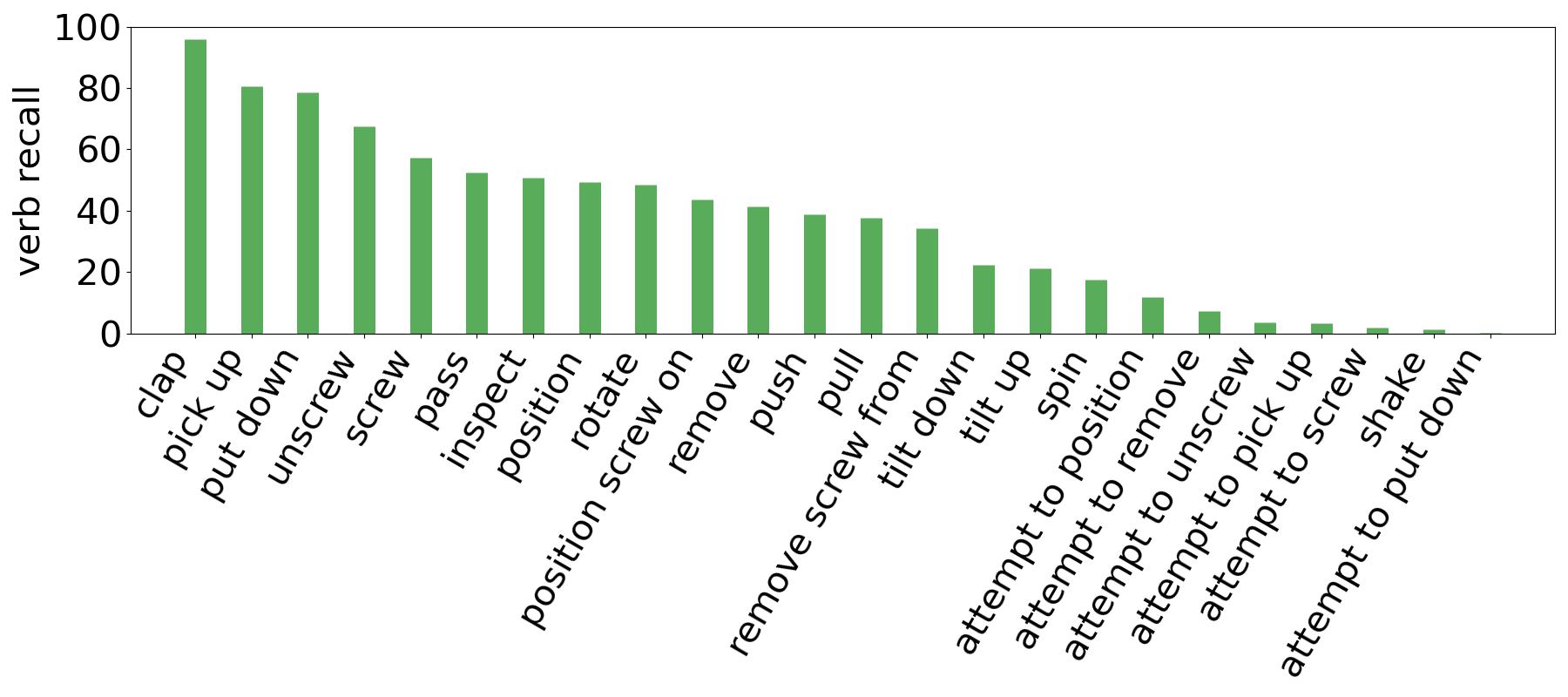} 
\caption{Action recognition object and verb recall.}
\label{fig:supp_verb_noun_recall}
\end{figure}

\subsection{Skill level} 
We did not observe a significant difference across skill levels for action recognition and anticipation tasks. A reason could be that those tasks are trained on fine-level labels while skill is more relevant for coarse actions. 

\subsection{Toy categories}
Figure~\ref{fig:toy_cats} shows the accuracy of action recognition and temporal action segmentation models for each toy category. The toy with the highest score is ``transporter''. Although we have only 4 toys in ``transporter'' category, there are 22 participants recording these toys. We think its high performance could be due to the large number of recordings. 
 
\subsection{Class-based evaluations}
\paragraph{Fine-grained actions.}
We present the recall of the objects and verbs for action recognition in Fig.~\ref{fig:supp_verb_noun_recall}. The verbs with the highest recall are ``clap'', ``pick up'' and ``put down'', while the tail verbs involving ``attempt to'' have the lowest recall. We also present the top 24 object classes in Fig.~\ref{fig:supp_verb_noun_recall}. It can be seen that enumerated wheels are among the top classes. 

\paragraph{Coarse actions.}
Based on the temporal action segmentation results, we further investigated the performance of verbs and objects. Out of 11 coarse verbs, the verbs with the highest recall are ``demonstrate'', ``attach'' and ``detach'', and the ones with the lowest recall are ``position'', ``remove'' and ``attempt to screw'', which are the tail verbs. The objects with the highest recall are ``chassis'' and ``interior'', which are the most common objects across toys. 

{\small
\bibliographystyle{ieee_fullname}
\bibliography{egbib}
}